\documentclass{article}

\usepackage{microtype}
\usepackage{graphicx}
\usepackage{subfigure}
\usepackage{booktabs} % for 
\usepackage{amsfonts,bm}

\usepackage{hyperref}
\usepackage[all]{hypcap}        % needed to help hyperlinks direct correctly;

\usepackage{amsmath}
\usepackage{amssymb}
\usepackage{mathtools}
\usepackage{amsthm}

\usepackage{url}
\usepackage{bbding}

\usepackage{array} % extended column commands
\usepackage{booktabs}
\usepackage{multirow}
\usepackage{tablefootnote}
\usepackage{footmisc}

\usepackage{graphicx}
\usepackage{wrapfig}
\usepackage{caption}
\usepackage{subcaption}

\usepackage{enumitem}
\usepackage{fnpos} % \makeFNbelow by default 

\newcommand{\sumedge}{\sum_{(i,j)\in\mathcal{E}} }
\newcommand{\sumedgehollow}{\sum_{(i,j)\in\mathcal{E}, i\neq j} }

\newcommand{\sumnode}{\sum_{i\in\mathcal{V}} }
\newcommand{\energy}{\mathcal{H}}
\newcommand{\Hhat}{\hat{\energy}}

\def\1{\bm{1}}

\def\vf{{\bm{f}}}

\def\vm{{\bm{m}}}

\def\vq{{\bm{q}}}

\def\vx{{\bm{x}}}
\def\vy{{\bm{y}}}
\def\vz{{\bm{z}}}

\def\mA{{\bm{A}}}

\def\mD{{\bm{D}}}

\def\mF{{\bm{F}}}

\def\mI{{\bm{I}}}

\def\mL{{\bm{L}}}
\def\mM{{\bm{M}}}

\def\mQ{{\bm{Q}}}

\def\mW{{\bm{W}}}

\def\mZ{{\bm{Z}}}

\def\gN{{\mathcal{N}}}

\def\gV{{\mathcal{V}}}

\newcommand{\cut}[1]{{}}

\newcommand{\vS}{{\mathbf{S}}}

\newcommand{\cE}{{\mathcal{E}}}
\newcommand{\cF}{{\mathcal{F}}}
\newcommand{\cG}{{\mathcal{G}}}

\newcommand{\cN}{{\mathcal{N}}}

\newcommand{\cV}{{\mathcal{V}}}

\newcommand{\RR}{\mathbb{R}}

\newcommand{\tr}{{\mathrm{tr}}} % trace
\makeatletter
\let\@@span\span
\def\sp@n{\@@span\omit\advance\@multicnt\m@ne}
\makeatother

\DeclareMathOperator*{\argmin}{arg\,min}

\newcommand{\bc}{\begin{center}}
\newcommand{\ec}{\end{center}}

\newcommand{\bdm}{\begin{displaymath}}
\newcommand{\edm}{\end{displaymath}}

\newcommand{\beq}{\begin{equation}}
\newcommand{\eeq}{\end{equation}}

\newcommand{\bfl}{\begin{flushleft}}
\newcommand{\efl}{\end{flushleft}}

\newcommand{\bt}{\begin{tabbing}}
\newcommand{\et}{\end{tabbing}}

\newcommand{\beqn}{\begin{align}}
\newcommand{\eeqn}{\end{align}}

\newcommand{\beqs}{\begin{align*}} % no equation numbers
\newcommand{\eeqs}{\end{align*}}  % no equation numbers

\newtheorem{remark}{Remark}

\DeclareMathAlphabet{\mathpzc}{OT1}{pzc}{m}{it}

\usepackage{natbib}

\newtheorem{theorem}{Theorem}
\newtheorem{lemma}{Lemma}

    \usepackage[final]{neurips_2024}

\usepackage[utf8]{inputenc} % allow utf-8 input
\usepackage[T1]{fontenc}    % use 8-bit T1 fonts
\usepackage{hyperref}       % hyperlinks
\usepackage{url}            % simple URL typesetting
\usepackage{booktabs}       % professional-quality tables
\usepackage{amsfonts}       % blackboard math symbols
\usepackage{nicefrac}       % compact symbols for 1/2, etc.
\usepackage{microtype}      % microtypography
\usepackage{xcolor}         % colors

\title{Robust Graph Neural Networks \\ via Unbiased Aggregation}

\author{
  Zhichao Hou\textsuperscript{\rm 1}\thanks{Equal contribution.} 
  \And
  Ruiqi Feng\textsuperscript{\rm 1}\footnotemark[1] 
  \And
  Tyler Derr\textsuperscript{\rm 2} 
  \And
  Xiaorui Liu\textsuperscript{\rm 1}\thanks{Corresponding author.}  
  \\
    \and
  \textsuperscript{\rm 1}North Carolina State University, 
  \textsuperscript{\rm 2}Vanderbilt University\\
  \and
  \texttt{\{zhou4,xliu96\}@ncsu.edu ruiqifeng.2024@gmail.com tyler.derr@vanderbilt.edu}\\
}

\begin{document}

\maketitle

\begin{abstract}

The adversarial robustness of Graph Neural Networks (GNNs) has been questioned due to the false sense of security uncovered by strong adaptive attacks despite the existence of numerous defenses.
In this work, we delve into the robustness analysis of representative robust GNNs and provide a unified robust estimation point of view to
understand their robustness and limitations.
Our novel analysis of estimation bias motivates the design of a 
robust and unbiased graph signal estimator. 
We then develop an efficient Quasi-Newton Iterative Reweighted Least Squares algorithm to solve the estimation problem, which is unfolded as robust unbiased aggregation layers in GNNs with theoretical guarantees.
Our comprehensive experiments confirm the strong robustness of our proposed model under various scenarios, and the ablation study provides a deep understanding of its advantages. Our code is available at \href{https://github.com/chris-hzc/RUNG}{https://github.com/chris-hzc/RUNG}. 

\end{abstract}

\section{Introduction}

\label{sec:intro}

Graph neural networks (GNNs) have gained tremendous popularity in recent years due to their ability to capture topological relationships in graph-structured data~\citep{ma2021deep}.
However, most GNNs are vulnerable to adversarial attacks, 
which can lead to a substantial decline in predictive performance
~\citep{zhang2020gnnguard,pro_gnn,softmedian}.
Despite the numerous defense strategies proposed to robustify GNNs, a recent study has revealed that most of these defenses are not as robust as initially claimed \citep{mujkanovic2022_are_defenses_for_gnns_robust}. 
Specifically, under adaptive attacks, they easily underperform the multi-layer perceptrons (MLPs) which do not utilize the graph topology information at all \citep{mujkanovic2022_are_defenses_for_gnns_robust}.
Therefore, it is imperative to thoroughly investigate the limitations of existing defenses and develop innovative robust GNNs to securely harness the topology information in graphs.

Existing defenses attempt to bolster the resilience of GNNs using diverse approaches.
For instance, Jaccard-GCN~\citep{wu2019adversarial} and SVD-GCN~\citep{svd_gcn} aim to denoise the graph by removing potential adversarial edges during the pre-processing procedure, while ProGNN~\citep{pro_gnn} learns the clean graph structure during the training process.
GRAND~\citep{grand} and robust training~\citep{deng2023batch, chen2020smoothing} also improve the training procedure through data augmentation.
GNNGuard~\citep{zhang2020gnnguard} and RGCN~\citep{zhu2019robust} reinforce their GNN architectures by heuristically reweighting edges in the graph. Additionally, there emerge some ODEs-inspired architectures including the GraphCON~\citep{rusch2022graph} and HANG~\citep{zhao2024adversarial} that demonstrate decent robustness.
Although most of these defenses exhibit decent robustness against transfer attacks, i.e., the attack is generated through surrogate models, they encounter
catastrophic performance drops when confronted with adaptive adversarial attacks that directly attack the victim model~\citep{mujkanovic2022_are_defenses_for_gnns_robust}. 
Concerned by the false sense of security, we provide a comprehensive study on existing defenses
under adaptive attacks.
Our preliminary study in Section~\ref{sec:bias} indicates that SoftMedian~\citep{softmedian}, TWIRLS~\citep{TWIRLS}, and ElasticGNN~\citep{elastic_gnn} exhibit closely aligned performance and notably outperform 
other defenses 
despite their apparent architectural differences. However, as attack budgets increase, these 
defenses still experience a severe performance decrease and underperform the graph-agnostic MLPs. These observations are intriguing, but the underlying reasons are still unclear. 

To unravel the aligned robustness and performance degradation of SoftMedian, TWIRLS, and ElasticGNN,
we delve into their theoretical understanding 
and unveil their inherent connections and limitations in the underlying principles. Specifically, their improved robustness can be understood from a unified view of $\ell_1$-based robust graph smoothing.
Moreover, we unearth the problematic estimation bias of $\ell_1$-based graph smoothing that allows the adversarial impact to accumulate as the attack budget escalates, which provides a plausible explanation of their declining robustness.
Motivated by these understandings, we propose a robust and unbiased graph signal estimator to reduce the estimation bias in GNNs. We design an efficient Quasi-Newton IRLS algorithm that unrolls as robust unbiased aggregation layers to safeguard GNNs against adversarial attacks. 
Our contributions can be summarized as follows:

\begin{itemize}
[leftmargin=0.2in]
\item We provide a unified view of $\ell_1$-based robust graph signal smoothing to justify the improved and closely aligned robustness of representative robust GNNs. Moreover, we reveal their estimation bias, which explains their severe performance degradation as the attack budgets increase. 

\item We propose a robust and unbiased graph signal estimator to mitigate the estimation bias in $\ell_1$-based graph signal smoothing and design an efficient Quasi-Newton IRLS algorithm to solve the non-smooth and non-convex estimation problem with theoretical guarantees.

\item The proposed algorithm can be readily unfolded as 
feature aggregation building blocks in GNNs, which not only provides clear interpretability but also covers many classic GNNs as special cases. 

\item Comprehensive experiments demonstrate that our proposed GNN significantly improves the robustness 
while maintaining clean accuracy. We also provide comprehensive ablation studies to validate its working mechanism. 
\end{itemize}

\section{Estimation Bias Analysis of Robust GNNs}
\label{sec:bias}

In Section~\ref{sec:robust-analysis}, we conduct a preliminary study to evaluate the robustness of several representative robust GNNs. In Section~\ref{sec:unified-view}, we establish a unified view as $\ell_1$-based models to uncover the inherent connections of three well-performing GNNs,
including SoftMedian, TWIRLS and ElasticGNN. In Section~\ref{sec:bias_analysis}, we leverage the bias of  $\ell_1$-based estimation to
explain the catastrophic performance degradation in the preliminary experiments.

\textbf{Notation.} 
Let $\cG=\{\cV, \cE\}$ be a graph with node set $\cV=\{v_1, \dots, v_n\}$ and edge set $\cE=\{e_1, \dots, e_m\}$. 
The adjacency matrix of $\cG$ is denoted as $\mA\in\{0,1\}^{n\times n}$ and the graph Laplacian matrix is $\mL=\mD-\mA$. $\mD=\text{diag} (d_1, \dots, d_n)$ is the degree matrix where $d_i = |\cN(i)|$ and  $\cN(i)$ is the neighborhood set of $v_i$.
The node feature matrix is denoted as $\mF = [\vf_1, \dots, \vf_n]^\top \in \mathbb{R}^{n\times d}$, and $\vf^{(0)}$ ($\mF^{(0)}$) denotes the node feature vector (matrix) before graph smoothing in decoupled GNN models.
Let $\Delta\in\{-1,0,1\}^{m\times n}$ be the incidence matrix whose $l$-th row denotes the $l$-th edge $e_l=(i,j)$ such that $\Delta_{li}=-1, \Delta_{lj}=1, \Delta_{lk} = 0 ~\forall k\notin \{i,j\}$.
$\tilde \Delta$ is its normalized version 
: $\tilde{\Delta}_{lj} = \Delta_{lj} / \sqrt{d_j}$.    
For a vector $\vx\in\RR^d$, we use $\ell_1$-based gragh smoothing penalty to denote either $\|\vx\|_{1}=\sum_{i} |\vx_{i}|$ or $\|\vx\|_{2}=\sqrt{\sum_i \vx_{i}^2}$. Note that we use $\ell_{2}$-based gragh smoothing penalty to denote $\|\vx\|_{2}^2 = \sum_i \vx_{i}^2$.

\subsection{Robustness Analysis}
\label{sec:robust-analysis}
 
To test the robustness of existing GNNs without the false sense of security, we perform a preliminary evaluation of existing robust GNNs against adaptive attacks.
We choose various baselines including the undefended MLP, GCN~\citep{gcn}, some of the most representative defenses 
in~\citep{mujkanovic2022_are_defenses_for_gnns_robust}, and two additional
robust models TWIRLS~\citep{TWIRLS} and ElasticGNN~\citep{elastic_gnn}.
We execute adaptive local evasion topological attacks and test the node classification accuracy on the Cora ML
and Citeseer datasets.
The detailed settings follow Section~\ref{sec:setting}.
From Figure~\ref{fig:preliminary_l1_models}, it can be observed that:

\begin{figure}[!ht]
    \centering
\includegraphics[width=0.7\textwidth]{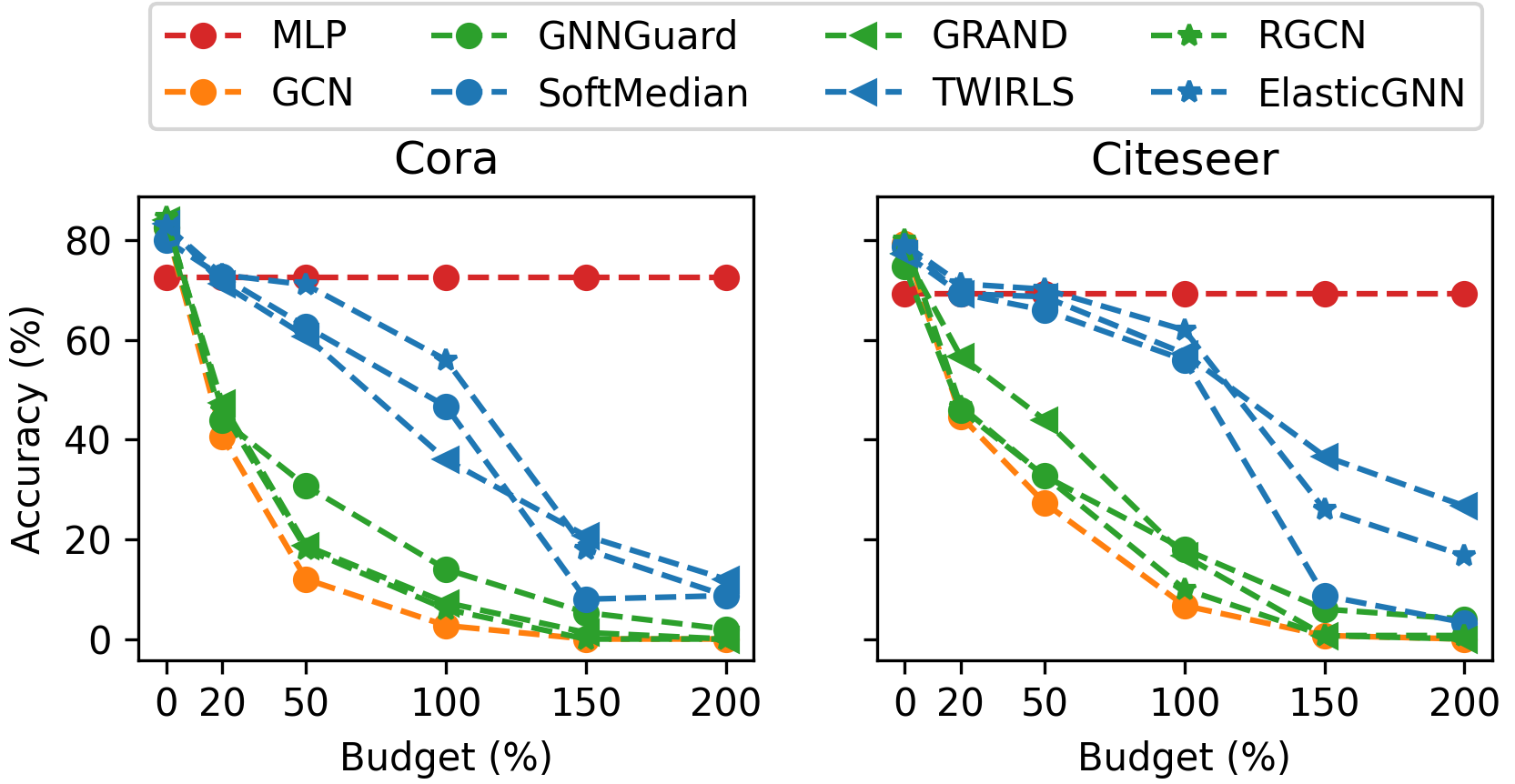}
    \caption{
    Robustness analysis under adaptive local attack. The perturbation budget ($x$-axis) is the number of edges allowed to be perturbed relative to the target node's degree. SoftMedian, TWIRLS, and ElasticGNN (blue curves) exhibit similarly aligned competitive robustness among all the selected robust GNNs, but all models experience catastrophic performance degradation as the attack budget increases.
    }
    \label{fig:preliminary_l1_models}
    \vspace{-0.18in}
\end{figure}

\begin{itemize}
[leftmargin=0.2in]
    \item  Among all the selected robust GNNs,
    only SoftMedian, TWIRLS, and ElasticGNN exhibit notable and closely aligned improvements in robustness whereas other GNNs do not show obvious improvement over undefended GCN. 
    \item SoftMedian, TWIRLS, and ElasticGNN encounter a similar catastrophic performance degradation as the attack budget scales up. Their accuracy easily drops below that of the graph-unware MLP, indicating their failure in safely exploiting the topology of the data.
\end{itemize}

\subsection{A Unified View of Robust Estimation}
\label{sec:unified-view}

Our preliminary study provides intriguing observations in Section~\ref{sec:robust-analysis}, but the underlying reasons behind these phenomena remain obscure.
This motivates us to delve into their theoretical understanding and explanation. In this section, we will compare the architectures of 
three well-performing GNNs, aiming to reveal their intrinsic connections.

\textit{SoftMedian}~\citep{softmedian} substitutes the GCN aggregation for enhanced robustness with the dimension-wise median $\vm_i\in\RR^d$ for all neighbors of each node $i\in\gV$. However, computing the median involves operations like ranking and selection, which is not compatible with the back-propagation training of GNNs. Therefore, the median is approximated as a differentiable weighted sum 
$\smash[b]{\tilde \vm_i= \frac{1}{Z}\sum_{j\in\gN(i)} w(\vf_j,\vm_i) \vf_j,\forall i\in\cV}$, where $\vm_i$ is the exact non-differentiable dimension-wise median, $\vf_j$ is the feature vector of the $j$-th neighbor, $w(\vx, \vy) = e^{-\beta\|\vx-\vy\|_2}$, and $Z=\sum_k w(\vf_k,\vm_k)$ is a normalization factor. 
In this way, the aggregation assigns the largest weights to the neighbors closest to the actual median.

\textit{TWIRLS}~\citep{TWIRLS}
utilizes the iteratively reweighted least squares (IRLS) algorithm to optimize the objective with parameter $\lambda$, and $\rho(y) = y$ is the default: 
\begin{equation}
\label{eq:TWIRLS}
2\lambda\sumedge\rho(\|\tilde{\vf_i} - \tilde{\vf_j}\|_2)+ \sumnode\|\tilde{\vf_i} - \tilde{\vf}^{(0)}\|_2^2, \tilde{\vf_i}
=(1 + \lambda d_i)^{-\frac{1}{2}}\vf_i.%\nonumber
\end{equation}

\textit{ElasticGNN}~\citep{elastic_gnn} proposes the elastic message passing 
% (EMP) 
which unfolds the proximal alternating predictor-corrector (PAPC) algorithm to minimize the objective with parameter $\lambda_{\{1,2\}}$:

\begin{equation}
\begin{split}
%\small
\label{eq:Elastic}
 &\frac{1}{2}\sumnode\|\vf_i - \vf^{(0)}_i\|_2^2+\lambda_1\sumedge\left\|\frac{\vf_i}{\sqrt{d_i}} - \frac{\vf_j}{\sqrt{d_j}}\right\|
_p + \lambda_2 \sumedge \left\|\frac{\vf_i}{\sqrt{d_i}} - \frac{\vf_j}{\sqrt{d_j}}\right\|_2^2, \text{where } p\in\{1, 2\}.\\
\end{split}
\end{equation}

\textbf{A Unified View of Robust Estimation.}
While these three approaches have seemingly different architectures, we provide a unified view of robust estimation to illuminate their inherent connections.
First, the objective of TWIRLS in Eq.~\eqref{eq:TWIRLS} can be considered as a particular case of ElasticGNN with $\lambda_2=0$ and $p=2$
when neglecting the difference in the node degree normalization.
However, TWIRLS and ElasticGNN leverage different optimization solvers, i.e., IRLS and PAPC, which leads to vastly different GNN layers.
%%%%
Second, SoftMedian approximates the computation of medians in a soft way of weighted sums, which can be regarded as approximately solving the dimension-wise median estimation problem~\citep{huber2004robust}: $\argmin_{\vf_i} \sum_{j\in\mathcal{N}(i)} \|\vf_i - \vf_{j}\|_1$. Therefore, SoftMedian can be regarded as the ElasticGNN with $\lambda_2=0$ and $p=1$.
% by setting $\lambda_1=0$ and $p=1$ in Eq.~\eqref{eq:Elastic}. 
We also note that the SoftMedoid~\citep{softmedoid} approach also resembles ElasticGNN with $\lambda_2=0$ and $p=2$, and the Total Variation GNN \citep{hansen2023tvgnn} also utilizes an $\ell_1$ estimator in spectral clustering.

The above analyses suggest that SoftMedian, TWIRLS, and ElasticGNN share the same underlying design principle of $\ell_1$-based robust graph signal estimation, i.e. a similar graph smoothing objective with edge difference penalties $\|\vf_i-\vf_j\|_1$ or $\|\vf_i-\vf_j\|_2$. However, they adopt different approximation solutions that result in distinct architecture designs. 
This unified view of robust estimation clearly explains their closely aligned performance. Besides, the superiority $\ell_1$-based models over the $\ell_2$-based models such as GCN~\citep{gcn},
whose graph smoothing objective is essentially $\sumedge \|\vf_i/\sqrt{d_i} - \vf_j/\sqrt{d_j}\|_2^2$~\citep{a_unifed_view_on_GNN_as_graph_denoising}, can 
% also
be 
% understood
explained
since $\ell_1$-based graph smoothing mitigates the impact of the outliers~\citep{elastic_gnn}.

\subsection{Bias Analysis and Performance Degradation}
\label{sec:bias_analysis}
The unified view of $\ell_1$-based graph smoothing we established in Section~\ref{sec:unified-view} not only explains their aligned robustness improvement but also provides a perspective to understand their failure as attack budgets scale up through an estimation bias analysis.

\begin{figure}[!ht]
    \vspace{-0.05in}
    \centering
    \includegraphics[width=1.0\textwidth]{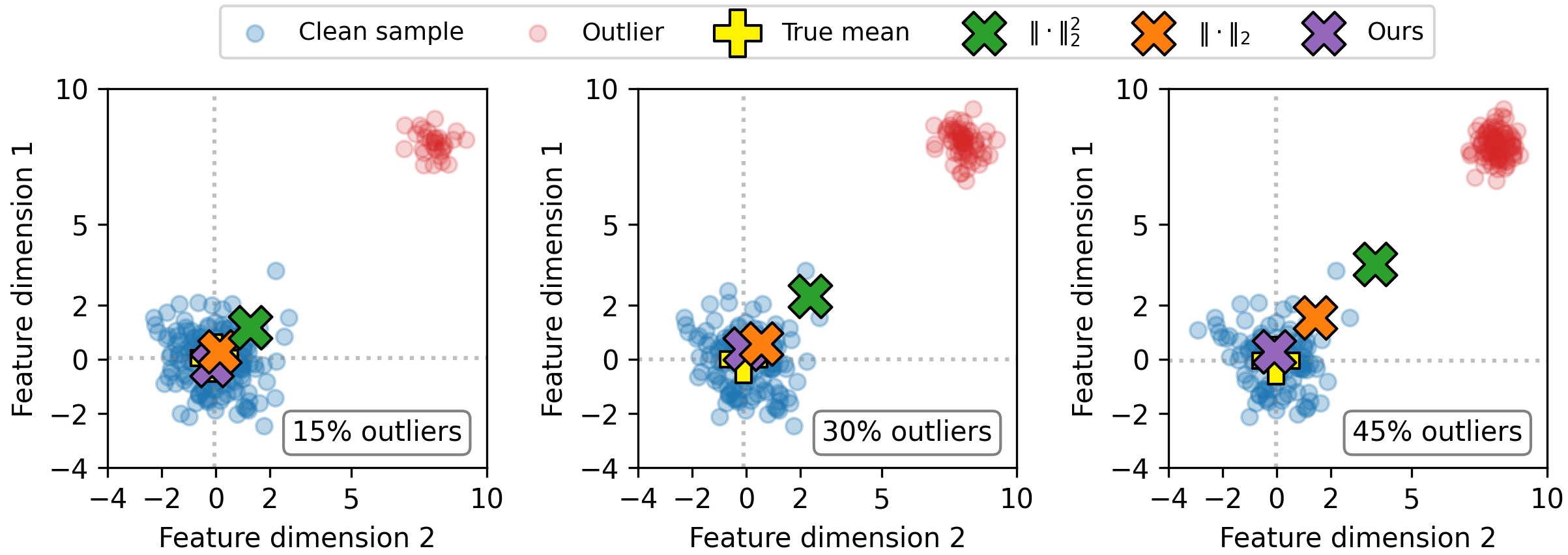}
    \vspace{-0.01in}
    \caption{
    Different mean estimators in the presence of outliers. The clean samples are the majority of data points following the Gaussian distribution $\mathcal{N}((0,0), 1 \cdot I)$, while the outliers are data points that deviate significantly from the main data pattern, following $\mathcal{N}((8,8), 0.5 \cdot I)$.
    $\ell_2$-estimator deviates far from the true mean, while the $\ell_1$-based estimator is more resistant to outliers. However, as the ratio of outliers escalates, the $\ell_1$-based estimator encounters a greater shift from the true mean, but
    our estimator still maintains a position close to the ground truth.
    }
    \label{fig:m_estimator_compare}
\vspace{-0.1in}
\end{figure}

\textbf{Bias of $\ell_1$-based Estimation.}
% Graph Smoothing.}
In the literature of high-dimensional statistics, it has been well understood that the $\ell_1$ regularization will induce an estimation bias. In the context of denoising~\citep{donoho1995noising} or variable selection~\citep{lasso_regression}, small coefficients $\beta$ are undesirable. To exclude small $\beta$ in the estimation, a soft-thresholding operator can be derived as $\vS_\lambda(\beta) = \text{sign}(\beta)\max(|\beta|-\lambda, 0)$. As a result, large $\beta$ are also shrunk by a constant, so the $\ell_1$ estimation is biased towards zero.
%%%%%

A similar bias effect also occurs in graph signal estimation in the presence of adversarial attacks.
For example, in TWIRLS (Eq.~\eqref{eq:TWIRLS}), %edge $e_k=(i,j)$ is reweighted by $w_{ij}=\|\tilde{\vf_i} - \tilde{\vf_j}\|_2^{-1}$. 
after the graph aggregation $\tilde\vf_i^{(k+1)} = \sum_{j\in\mathcal{N}(i)}w_{ij}\tilde\vf_j^{(k)}$ where $w_{ij}=\|\tilde{\vf_i} - \tilde{\vf_j}\|_2^{-1}$, %edge $e_k = (i,j)$ will shrink 
the edge difference $\tilde\vf_i - \tilde\vf_j$ will shrink towards zero.
%by the unit vector $\vu_{\tilde\vf_i - \tilde\vf_j}$.
Consequently, every 
adversarial edge the attacker adds will induce a bias that can be accumulated and amplified when the attack budget scales up.

\textbf{Numerical Simulation.}
% To better illustrate and validate our statement, 
To provide a more intuitive illustration of the estimation bias of $\ell_1$-based models, we simulate a mean estimation problem on synthetic data since most message passing schemes in GNNs essentially estimate the mean of neighboring node features. The results in Figure~\ref{fig:m_estimator_compare} shows that $\ell_1$-based estimator is more resistant than $\ell_2$-based estimator. However, as the ratio of outliers escalates, the $\ell_1$-based estimator encounters a greater shift from the true mean due to the accumulated bias caused by outliers.
This observation explains why $\ell_1$-based graph smoothing models suffer from catastrophic degradation under large attack budgets.  The detailed simulation settings and results are available in Appendix~\ref{sec:bias_details}.

% \newpage

\section{Robust GNNs with Unbiased Aggregation}\label{sec:algo}

In this section, we first design a robust unbiased estimator to reduce the bias in graph signal estimation in Section~\ref{sec:robust_graph_estimator}
and propose an efficient second-order IRLS
% iteratively reweighted least squares 
algorithm to compute the robust estimator with theoretical convergence guarantees in Section~\ref{sec:qn-irls}. Finally, we unroll the proposed algorithm as the robust unbiased feature aggregation layers in GNNs in Section~\ref{sec:robust-graph-aggregation}.

\subsection{Robust and Unbiased Graph Signal Estimator}
\label{sec:robust_graph_estimator}

Our study and analysis in Section~\ref{sec:bias} have shown that while $\ell_1$-based methods outperform $\ell_2$-based methods in robustness, 
% graph smoothing 
they still suffer from the accumulated estimation bias, leading to severe performance degradation under large perturbation budgets. 
This motivates us to design a robust and unbiased graph signal estimator that derives unbiased robust aggregation for GNNs with stronger resilience to attacks. % of large budgets.

Theoretically, the estimation bias in Lasso regression has been discovered and analyzed in high-dimensional statistics~\citep{adaptive_lasso}. 
Statisticians have proposed adaptive Lasso~\citep{adaptive_lasso} and many non-convex penalties such as Smoothly Clipped Absolute Deviation (SCAD)~\citep{scad_regression} and Minimax Concave Penalty (MCP)~\citep{mcp_regression} to alleviate this bias.
Motivated by these advancements,
% (MCP in particular), 
we propose a Robust and Unbiased Graph signal Estimator (RUGE) as follows:
% \begin{small}
% \vspace{-0.2in}
\begin{equation}\label{eq:mcp_obj}
\argmin_{\mF} 
\energy (\mF)= \sumedge  \rho_\gamma(\left\|\frac{\vf_i}{\sqrt{d_i}} - \frac{\vf_j}{\sqrt{d_j}}\right\|_2) + \lambda \sumnode \|\vf_i - \vf_i^{(0)}\|_2^2,
%&+ \lambda \sumnode \|\vf_i - \vf_i^{(0)}\|_2^2, 
\end{equation}
% \end{small}
where $\rho_\gamma(y)$ 
denotes the function that penalizes the feature differences on edges by 
% the non-convex 
MCP: 
\begin{equation}
    \rho_\gamma(y) =     
    \begin{cases}
      y - \frac{y^2}{2\gamma} & \text{if } y < \gamma\\
      \frac{\gamma}{2} & \text{if } y \ge \gamma  
    \end{cases}.
\end{equation}

\begin{wrapfigure}{r}{0.3\textwidth}
    \centering
    \vspace{-0.2in}
    % \hspace{-0.2in}
    \includegraphics[width=0.3
\textwidth]{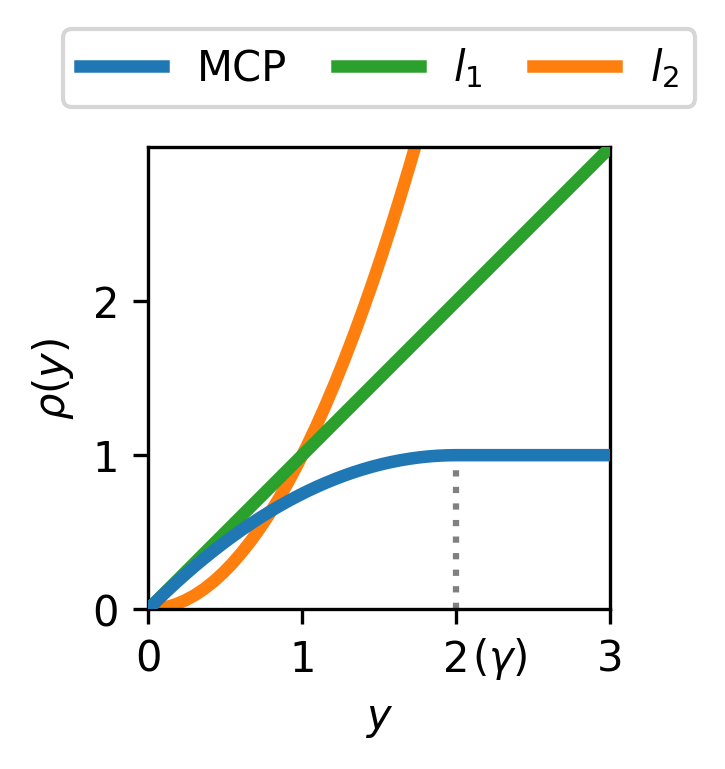}
    % \caption{Different smoothing penalties.}
    % \vspace{-0.15in}
    % \vspace{-0.2in}
    \caption{ Penalties.
    }
    \label{fig:att_demo_rho}
    % \vspace{-0.2in}
\end{wrapfigure}
As shown in Figure~\ref{fig:att_demo_rho}, MCP closely approximates the $\ell_1$ norm when $y$ is small since the quadratic term $\frac{y^2}{2\gamma}$ is negligible, and it becomes a constant value when $y$ is large. This transition can be adjusted by the thresholding parameter $\gamma$.
When $\gamma$ approaches infinity, the penalty $\rho_\gamma(y)$ reduces to the $\ell_1$ norm. Conversely, when $\gamma$ is very small, the ``valley'' of $\rho_\gamma$ near zero is exceptionally sharp, so $\rho_\gamma(y)$ approaches the $\ell_0$ norm and becomes a constant for a slightly larger $y$.
This enables RUGE to 
suppress smoothing on edges whose node differences exceeding the threshold $\gamma$. 
This not only mitigates the estimation bias against outliers but also maintains the estimation accuracy in the absence of outliers. 
The simulation in Figure~\ref{fig:m_estimator_compare} verifies that our proposed estimator ($\eta(\vx) \coloneqq \rho_{\gamma}(\|\vx\|_2)$) can recover the true mean despite the increasing outlier ratio when the outlier ratio is below the theoretical optimal breakdown point.

\subsection{Quasi-Newton IRLS}
\label{sec:qn-irls}

Despite the advantages discussed above, the proposed RUGE in Eq.~\eqref{eq:mcp_obj} 
% involving MCP 
is non-smooth and non-convex, which results in challenges for deriving efficient numerical solutions that can be readily unfolded as neural network layers.
In the literature, researchers have developed optimization algorithms for MCP-related problems,
% several sophisticated algorithms have been proposed to optimize 
% related objectives, 
such as the Alternating Direction Multiplier Method (ADMM) and Newton-type algorithms~\citep{scad_regression, mcp_regression, varma2019vector}.
However, due to their excessive computation and memory requirements as well as the incompatibility with back-propagation training, these algorithms are not well-suited for the construction of feature aggregation layers in GNNs.
To solve these challenges, we propose an efficient Quasi-Newton  Iteratively Reweighted Least Squares algorithm (QN-IRLS) to solve the estimation problem in Eq.~\eqref{eq:mcp_obj}.

\textbf{IRLS.} Before stepping into our QN-IRLS, we first introduce the main idea of iteratively reweighted least squares (IRLS)~\citep{holland1977robust}  and analyze its weakness in convergence. 
% In fact, 
IRLS aims to circumvent the 
% intractable 
non-smooth $\smash[b]{\energy(\mF)}$ in Eq.~\eqref{eq:mcp_obj} by computing its quadratic upper bound $\smash[b]{\Hhat}$ based on $\smash[b]{\mF^{(k)}}$ in each iteration $k$ and optimizing $\smash[b]{\Hhat}(\cF)$: 
% \begin{small}
% \vspace{-0.2in}
\begin{equation}\label{eq:Hhat}
    \Hhat(\mF) =
    %\sumedgehollow 
    \sum_{(i,j)\in \mathcal{E}}
    \mW_{ij}^{(k)}\left\|\frac{\vf_i}{\sqrt{d_i}} -\frac{\vf_j}{\sqrt{\smash[b]{d_j}}}\right\|_2^2 + \lambda\sumnode \|\vf_i - \vf_i^{(0)}\|_2^2,
\end{equation}
% \end{small}

where 
$\mW^{(k)}_{ij} = \1_{i\neq j}\frac{d \rho_\gamma(y_{ij})}{d y_{ij}^2}\big|_{y_{ij}=y_{ij}^{(k)}}$\footnote{$\mW_{ij}$ is defined as $\frac{d\rho(y)}{d y^2}\big|_{y=y_{ij}^{(k)}}$ so that the quadratic upper bound $\Hhat$ is tight at $\mF^{(k)}$ according to \autoref{lemma:quadratic_hessian_upper_bound}. 
The diagonal terms of $\mW$ are set to zero to avoid undefined derivative of $\smash[b]{\frac{d\rho(y)}{dy^2}\big|_{y=0}}$ as discussed in Remark~\ref{remark:hollow_W}.}
, 
%$\smash[b]{y^{(k)}_{ij} = \big\|\frac{\vf_i^{(k)}}{\sqrt{d_i}} - \frac{\vf_j^{(k)}}{\sqrt{\smash[b]{d_j}}}\big\|_2}$. 
where $\smash[b]{y^{(k)}_{ij} = \big\|\vf_i^{(k)}/\sqrt{d_i} - \vf_j^{(k)}/\sqrt{\smash[b]{d_j}}\big\|_2}$ and $\rho_\gamma(\cdot)$ is the MCP function. 
For the detailed proof of the upper bound, please refer to Lemma~\ref{lemma:weighted_majorizer} in Appendix~\ref{sec:convergence}.
Then, each iterative step of IRLS can be formulated as the first-order gradient descent 
% iteration 
for $\Hhat(\mF)$:
% \begin{small}
%\small{
\begin{equation}
\label{eq:rw_update_f1}
\begin{split}
&\mF^{(k+1)} = \mF^{(k)} - \eta \nabla \Hhat(\mF^{(k)}) 
% \nonumber \\
% &
% = \mF^{(k)} - \eta \left( 2 (\text{diag}(\vq^{(k)}) - \mW^{(k)}\odot\tilde{\mA}) \mF^{(k)} + 2 \lambda \mF^{(k)} - 2 \lambda \mF^{(0)} \right),
= \mF^{(k)} - \eta \left( (\hat \mQ^{(k)} - 2\mW^{(k)}\odot\tilde{\mA}) \mF^{(k)}-2\lambda \mF^{(0)} \right),
\end{split}
\end{equation}
% \end{small}
where $\eta$ is the 
% update 
step size, $\hat \mQ^{(k)} = 2(\text{diag}(\vq^{(k)}) + \lambda \mI)$, and $\vq^{(k)}_m = \sum_{j} \mW^{(k)}_{mj}\mA_{mj} / d_m$.
% $\tilde{\mA}_{ij} = \frac{\mA_{ij}}{\sqrt{d_id_j}}$.
% $\tilde{\mA}_{ij} = \mA_{ij}/\sqrt{d_id_j}$ 
% The convergence condition of Eq.~\eqref{eq:rw_update_f1} 
Its convergence condition is given in Theorem~\ref{theorem:gd_descent},
% and the detailed proof is 
% whose proof is presented in Appendix~\ref{sec:convergence}. 
with a proof in Appendix~\ref{sec:convergence}.

{
\renewcommand{\thetheorem}{\ref{theorem:gd_descent}}
\begin{theorem} 
If $\mF^{(k)}$ follows the update rule in Eq.~\eqref{eq:rw_update_f1} where $\rho$ defining $\mW$ satisfies that $\smash[b]{{\frac{d\rho(y)}{dy^2}}}$ is non-decreasing $\forall y\in(0,\infty)$, 
then a sufficient condition for $\energy(\mF^{(k+1)})\le \energy(\mF^{(k)})$ is that the step size $\eta$ satisfies $0 < \eta \le \|\text{diag}(\vq^{(k)}) - \mW^{(k)}\odot\tilde{\mA} + \lambda\mI\|_2^{-1}.$
% \begin{equation}
%      0 < \eta \le \|\text{diag}(\vq) - \mW^{(k)}\odot\tilde{\mA} + \lambda\mI\|_2^{-1}.
% \end{equation}    
    \end{theorem}
\addtocounter{theorem}{-1}
}

\textbf{Quasi-Newton IRLS.}
Theorem~\ref{theorem:gd_descent} suggests the difficulty in the proper selection of stepsize for (first-order) IRLS due to its non-trivial dependency on the graph ($\tilde\mA$) and the dynamic terms ($\vq^{(k)}$ and $\mW^{(k)}$)
\footnote{A related work~\citep{TWIRLS} adopts IRLS algorithm
% ~\citep{holland1977robust}
to optimize the problem in Eq.~\eqref{eq:TWIRLS}. A preconditioned version is proposed to handle the unnormalized graph Laplacian, but its step size needs to satisfy $\eta \leq  \|\Delta^\top{\bf \Gamma}^{(k)}\Delta+\lambda\mI\|_2^{-1}$ as shown in Lemma 3.3 of ~\citep{TWIRLS}, which is expensive to estimate.
}.
The dilemma is that a small stepsize will lead to slow convergence but a large step easily causes divergence and instability as verified by our experiments in Section~\ref{sec:ablation}  (Figrue~\ref{fig:convergence_of_ours}), which reveals its critical shortcoming for the construction of GNN layers.
% reliable GNN aggregation layers.

To overcome this limitation, we aim to propose a second-order Newton method, 
$\mF^{(k+1)} =\mF^{(k)}- (\nabla^2\Hhat(\mF^{(k)}))^{-1} \nabla \Hhat(\mF^{(k)})$, 
% $$\mF^{(k+1)} =\mF^{(k)}- (\nabla^2\Hhat(\mF^{(k)}))^{-1} \nabla \Hhat(\mF^{(k)}),$$ 
to achieve faster convergence and stepsize-free hyperparameter tuning by better capturing the geometry of the optimization landscape. However, obtaining the analytic expression for the inverse Hessian matrix $(\nabla^2\Hhat(\mF^{(k)}))^{-1}\in\RR^{n\times n}$ is intractable, and the numerical solution requires expensive computation for large graphs. 
% To resolve this challenge,
Therefore, we propose a novel Quasi-Newton IRLS algorithm (QN-IRLS) 
that approximates the Hessian matrix $\nabla^2\Hhat(\mF^{(k)})=2(\text{diag}(\vq^{(k)}) - \mW^{(k)}\odot\tilde\mA + \lambda\mI)$
by the diagonal matrix $\hat{\mQ}^{(k)}=2(\text{diag}(\vq^{(k)}) + \lambda\mI)$ such that the inverse is trivial. 
The proposed QN-IRLS works as follows:
\begin{equation}
\label{eq:rw_update_f2}
\begin{split}
    &\mF^{(k+1)}=\mF^{(k)} - \big(\hat{\mQ}^{(k)}\big)^{-1} \nabla \Hhat(\mF^{(k)})=
    (\text{diag}(\vq^{(k)}) + \lambda \mI)^{-1}\left( ( \mW^{(k)}\odot\tilde{\mA}) \mF^{(k)} +   \lambda \mF^{(0)}\right),
    % \vspace{-5pt}
\end{split}
\end{equation}
where $(\hat{\mQ}^{(k)})^{-1}$ automatically adjusts the per-coordinate stepsize according to the local geometry of the optimization landscape, $\vq^{(k)}$ and $\mW^{(k)}$ are defined as in Eq. \eqref{eq:Hhat} and \eqref{eq:rw_update_f1}.
In this way,
QN-IRLS provides faster convergence without needing to select a stepsize. The convergence is guaranteed by Theorem~\ref{theorem:preconditioned_descent} with the proof in \autoref{sec:convergence}.

{
\renewcommand{\thetheorem}{\ref{theorem:preconditioned_descent}}
    \begin{theorem}
        If $\mF^{(k+1)}$ follows update rule in Eq.~\eqref{eq:rw_update_f2}, where $\rho$ satisfies that $\frac{d\rho(y)}{dy^2}$ is non-decreasing $\forall y\in(0,\infty)$,
        it is guaranteed that $\energy(\mF^{(k+1)})\le \energy(\mF^{(k)})$. 
    \end{theorem}
\addtocounter{theorem}{-1}
}

\subsection{GNN with Robust Unbiased Aggregation}
\label{sec:robust-graph-aggregation}

\begin{wrapfigure}{r}{0.28\textwidth}
\centering
\vspace{-0.18in}
\includegraphics[width=0.27
\textwidth]{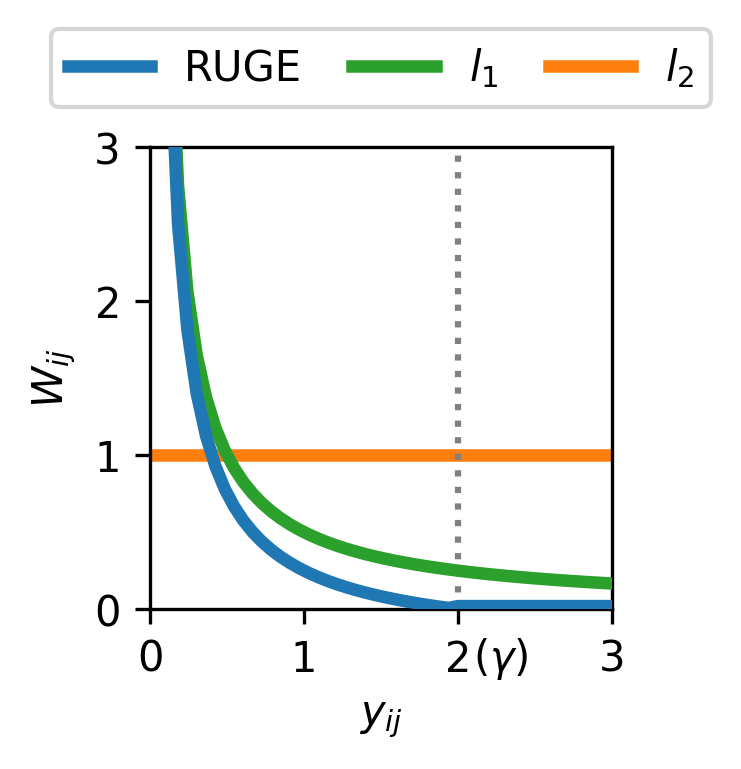}
% \caption{Different smoothing penalties.}
\vspace{-0.1in}
\caption{$\frac{d\rho(y)}{d y^2}$.
}
\label{fig:att_demo_drho}
\vspace{-0.35in}
\end{wrapfigure}

The proposed QN-IRLS provides an efficient algorithm
% message-passing scheme 
to optimize the RUGE in Eq.~\eqref{eq:mcp_obj} with a theoretical convergence guarantee.
Instantiated with $\rho=\rho_\gamma$, each iteration in QN-IRLS in Eq.~\eqref{eq:rw_update_f2} can be used as one layer in GNNs, which yields the \underline{R}obust \underline{Un}biased A\underline{g}gregation (RUNG): 
% \begin{small}
% \vspace{-0.2in}
\begin{equation}
\begin{split}
\label{eq:rw_update_f3}
&\mF^{(k+1)}=  (\text{diag}(\vq^{(k)}) + \lambda \mI)^{-1}\left( ( \mW^{(k)}\odot\tilde{\mA}) \mF^{(k)} +   \lambda \mF^{(0)}\right),
\end{split}
\end{equation}
% \end{small}
where $\vq^{(k)}_m = \sum_{j} \mW^{(k)}_{mj}\mA_{mj} / d_m$ as in Eq.~\eqref{eq:rw_update_f1}
, $\mW^{(k)}_{ij} = \1_{i\neq j} \max(0, \frac{1}{\smash[b]{y_{ij}^{(k)}}} - \frac{1}{\gamma})$ and $\smash[b]{y^{(k)}_{ij} = \big\|\frac{\vf_i^{(k)}}{\sqrt{d_i}} - \frac{\vf_j^{(k)}}{\sqrt{\smash[b]{d_j}}}\big\|_2}$.

\textbf{Interpretability.} 
The proposed RUNG can be interpreted intuitively with edge reweighting. In Eq.~\eqref{eq:rw_update_f3}, the normalized adjacency matrix $\tilde{\mA}$ is reweighted by $\mW^{(k)}$, where $\mW^{(k)}_{ij} = \frac{d \rho(y)}{d y^2}|_{y=y_{ij}^{(k)}}$. 
It is shown in \autoref{fig:att_demo_drho} that $\mW_{ij}$ becomes zero for any edge $e_{k}=(i,j)$ with a node difference $y_{ij}^{(k)} \ge \gamma$, thus pruning suspicious edges. This implies RUNG's strong robustness under large-budget adversarial attacks. With the inclusion of the skip connection $\mF^{(0)}$, $\text{diag}(\vq^{(k)}) + \lambda \mI$ can be seen as a normalizer of the layer output.

\textbf{Relations with Existing GNNs.}
RUNG can adopt different $\rho$ that \autoref{theorem:preconditioned_descent} allows, thus covering many classic GNNs as special cases.
When $\rho(y) = y^2$, RUNG 
in Eq.~\eqref{eq:rw_update_f3} 
exactly reduces to APPNP~\citep{appnp} 
% i.e., 
($\mF^{(k+1)} = \frac{1}{1+\lambda}\tilde\mA\mF^{(k)} + \frac{\lambda}{1+\lambda}\mF^{(0)}$) and GCN
% ~\citep{gcn}, 
% i.e., 
($\mF^{(k+1)} = \tilde\mA\mF^{(k)}$)
% can be acquired by 
if chosing $\lambda=0$. 
When $\rho(y) = y$, the objective of RUGE is equivalent to ElasticGNN with $p=2$, which is analogous to SoftMedian and TWIRLS due to their inherent connections as $\ell_1$-based graph smoothing.

\textbf{Complexity analysis.} RUNG is scalable with time complexity of $O(k(m + n)d)$ and space complexity $O(m+nd)$, where $m$ is the number of edges, $d$ is the number of features, $n$ is the number of nodes, and $k$ is the number of GNN layers. Therefore, the complexity of our RUNG is comparable to normal GCN (with a constant difference) and it is feasible to implement. The detailed discussions about computation efficiency can be found in Appendix~\ref{sec:compute_efficiency}.

\section{Experiment}\label{sec:exp}

In this section, we perform comprehensive experiments to validate the robustness of the proposed RUNG. Besides, ablation studies show 
% the convergence and defense mechanism of RUNG.
its convergence and defense mechanism.

\subsection{Experiment Setting}
\label{sec:setting}

\textbf{Datasets.}
We test our RUNG with the node classification task on two widely used real-world citation networks, Cora ML and Citeseer~\citep{cora_citeseer}, as well as a large-scale networks Ogbn-Arxiv~\citep{hu2020open}.
We adopt the data split of $10\%$ training, $10\%$ validation, and $80\%$ testing, and report the classification accuracy of the attacked nodes following \citep{mujkanovic2022_are_defenses_for_gnns_robust}. Each experiment is averaged over $5$ different random splits.

\textbf{Baselines.}
To evaluate the performance of RUNG, we compare it to $\ell_2$ other representative baselines. Among them, MLP, GCN~\citep{gcn}, APPNP~\citep{appnp}, and GAT~\citep{velickovic2017graph} are undefended vanilla models. GNNGuard~\citep{zhang2020gnnguard}, RGCN~\citep{zhu2019robust}, GRAND~\citep{grand}, ProGNN~\citep{pro_gnn}, Jaccard-GCN~\citep{wu2019adversarial}, SVD-GCN~\citep{svd_gcn}, EvenNet~\citep{lei2022evennet}, HANG~\citep{zhao2024adversarial}, NoisyGNN~\citep{ennadir2024simple}, and GARNET~\citep{deng2022garnet} are representative robust GNNs. Besides, 
SoftMedian and TWIRLS are representative methods with $\ell_1$-based graph smoothing
\footnote{We do not include ElasticGNN because it is still unclear how to attack it adaptively due to its special incident matrix formulation~\citep{elastic_gnn}. In the preliminary study (Section~\ref{sec:robust-analysis}), we evaluate the robustness of ElasticGNN following the unit test setting proposed in~\citep{mujkanovic2022_are_defenses_for_gnns_robust}.}
We also evaluate a variant of TWIRLS 
with thresholding attention (TWIRLS-T).
For RUNG, we test two variants: default RUNG (Eq.~\eqref{eq:rw_update_f3}) and RUNG-$\ell_1$ with $\ell_1$ penalty ($\rho(y)=y$).

\textbf{Hyperparameters.}
The model hyperparameters including
learning rate, weight decay, and dropout rate are tuned as in \citep{mujkanovic2022_are_defenses_for_gnns_robust}. Other hyperparameters follow the settings in the original papers. 
RUNG uses an MLP connected to 10 graph aggregation layers following the decoupled GNN architecture of APPNP. $\smash[b]{\hat\lambda=\frac{1}{1+\lambda}}$ is tuned in \{0.7, 0.8, 0.9\}, and $\gamma$ tuned in $\{0.5,1,2,3,5\}$. We chose the hyperparameter setting that yields the best robustness without a notable impact (smaller than $1\%$) on the clean accuracy following~\citep{Bojchevski2019CertifiableRT}.

\textbf{Attack setting.}
We use the 
% projected gradient descent (PGD) 
PGD attack~\citep{xu2019topology} to execute the adaptive \emph{evasion} and \emph{poisoning} topology attack since it delivers the strongest attack in most settings \citep{mujkanovic2022_are_defenses_for_gnns_robust}. The adaptive attack setting is provided in Appendix~\ref{sec:atk_settings}.
The adversarial attacks aim to misclassify specific target nodes (\emph{local attack}) or the entire set of test nodes (\emph{global attack}). 
To avoid a false sense of robustness, our \emph{adaptive} attacks directly target the victim model instead of the surrogate model. Additionally, we include the \emph{transfer} attacks with a 2-layer GCN as the surrogate model. We also include graph injection attack following the setting in TDGIA~\citep{zou2021tdgia}.

\subsection{Adversarial Robustness}\label{sec:exp_main}

\begin{table}[!ht]
\centering
% \tiny
\caption{Adaptive local attack on Cora ML. The \textbf{best} and \underline{second} are marked.}
\resizebox{1.0\textwidth}{!}{%
    \begin{tabular}{l c c c c c c}
        \toprule
        \multirow{1}{*}{Model} & \multirow{1}{*}{0\% (Clean)} & \multicolumn{1}{c}{$20\%$} & \multicolumn{1}{c}{$50\%$} & \multicolumn{1}{c}{$100\%$} & \multicolumn{1}{c}{$150\%$} & \multicolumn{1}{c}{$200\%$} \\
        \midrule
        MLP
        & $72.6 \pm 6.4$ & $72.6 \pm 6.4$ & \underline{$72.6 \pm 6.4$} & $\mathbf{72.6 \pm 6.4}$ & $\mathbf{72.6 \pm 6.4}$& $\mathbf{72.6 \pm 6.4}$ \\
        GCN  &$ 82.7 \pm 4.9$ & $40.7 \pm 10.2$  & $12.0 \pm 6.2$  & $2.7 \pm 2.5$  & $0.0 \pm 0.0$ & $0.0 \pm 0.0$   \\
        APPNP & $\mathbf{84.7 \pm 6.8}$ & $50.0 \pm 13.0$  & $27.3 \pm 6.5$  & $14.0 \pm 5.3$  & $3.3 \pm 3.0$ & $0.7 \pm 1.3$ \\
        GAT   & $80.7 \pm 10.0$  & $30.7 \pm 16.1$  & $16.0 \pm 12.2$  & $11.3 \pm 4.5$  & $1.3 \pm 1.6$  & $2.0 \pm 1.6$\\
        \midrule
        GNNGuard & $82.7 \pm 6.7 $ & $44.0 \pm 11.6$  & $30.7 \pm 11.6$  & $14.0 \pm 6.8$  & $5.3 \pm 3.4$ & $2.0 \pm 2.7$ \\
        RGCN   & \underline{$84.6 \pm 4.0$} & $46.0 \pm 9.3$  & $18.0 \pm 8.1$  & $6.0 \pm 3.9$  & $0.0 \pm 0.0$ & $0.0 \pm 0.0$    \\
        GRAND  & $84.0 \pm 6.8$ & $47.3 \pm 9.0$  & $18.7 \pm 9.1$  & $7.3 \pm 4.9$  & $1.3\pm1.6$ & $0.0 \pm 0.0$    \\
        ProGNN  & $\mathbf{84.7 \pm 6.2}$  & $47.3 \pm 10.4$  & $21.3 \pm 7.8$  & $4.0 \pm 2.5$  & $0.0 \pm 0.0$  & $0.0 \pm 0.0$ \\
        Jaccard-GCN   & $81.3 \pm 5.0$  & $46.0 \pm 6.8$  & $17.3 \pm 4.9$  & $4.7 \pm 3.4$  & $0.7 \pm 1.3$  & $0.7 \pm 1.3$  \\
        GARNET&$ 82.4\pm6.8 $  &$ 70.9\pm7.5 $  &$ 61.9\pm7.9 $  &$ 42.7\pm9.3 $  &$ 11.6\pm3.4 $  &$ 9.6\pm3.5 $  \\
        HANG&$ 83.1\pm 7.4 $  &$ 71.2\pm 7.8 $  &$ 60.1\pm 6.3 $  &$ 39.5\pm3.4 $  &$ 9.4\pm2.3 $  &$5.6 \pm2.3 $  \\
        NoisyGNN&$ 82.5\pm 5.6 $  &$ 57.4\pm 5.7 $  &$ 47.8\pm 6.2 $  &$ 36.1\pm 4.5 $  &$ 5.8\pm 3.4 $  &$ 4.1\pm 1.2 $  \\
        EvenNet&$ 83.4\pm 8.1 $  &$64.8 \pm 6.9$  &$ 56.1\pm 5.6 $  &$29.5 \pm 5.3 $  &$ 3.1\pm 1.1 $  &$ 1.1\pm 1.3 $  \\
        GraphCON&$81.3\pm 7.1 $&$67.3\pm7.3 $&$54.7\pm 7.1 $&$41.3\pm 6.2 $&$4.1\pm 2.1 $&$2.3\pm1.2 $\\
        SoftMedian  & $80.0 \pm 10.2$ & $\underline{72.7 \pm 13.7}$  & $62.7 \pm 12.7$  & $46.7 \pm 11.0$  & $8.0\pm4.5$ & $8.7 \pm 3.4$ \\ 
        TWIRLS & $83.3 \pm 7.3$ & $71.3 \pm 8.6$  & $60.7 \pm 11.0$  & $36.0 \pm 8.8$  & $20.7\pm10.4$ & $12.0 \pm 6.9$   \\
        TWIRLS-T  & $82.0 \pm 4.5$  & $70.7 \pm 4.4$  & $62.7 \pm 7.4$  & $54.7 \pm 6.2$  & $44.0 \pm 11.2$  & $40.7 \pm 11.8$ \\ 
        \midrule
        RUNG-$\ell_1$ (Ours) & $84.0 \pm 6.8$  & $\underline{72.7 \pm 7.1}$  & $62.7 \pm 11.2$  & $53.3 \pm 8.2$  & $22.0 \pm 9.3$  & $14.0 \pm 7.4$  \\ 
        RUNG (Ours) & $84.0 \pm 5.3$  & $\mathbf{75.3 \pm 6.9}$  & $\mathbf{72.7 \pm 8.5}$  & \underline{$70.7 \pm 10.6$}  & \underline{$69.3 \pm 9.8$}  & \underline{$69.3 \pm 9.0$} \\ 
        \bottomrule
    \end{tabular}
    }
    \label{cora_adaptive_local_evasion} 
\end{table}

\begin{table}[!ht]
\centering
\caption{Adaptive global attack on Cora ML. The \textbf{best} and \underline{second} are marked.
}
\resizebox{1.0\textwidth}{!}{%
    \begin{tabular}{l c c c c c c}
        \toprule
        \multirow{1}{*}{Model} & \multirow{1}{*}{0\% (Clean)} & \multicolumn{1}{c}{$5\%$} & \multicolumn{1}{c}{$10\%$} & \multicolumn{1}{c}{$20\%$} & \multicolumn{1}{c}{$30\%$} & \multicolumn{1}{c}{$40\%$} \\
        \midrule
        MLP
        & $65.0 \pm 1.0$ & $65.0 \pm 1.0$ & $65.0 \pm 1.0$ & $65.0 \pm 1.0$ & $\underline{65.0 \pm 1.0}$ & \underline{$65.0 \pm 1.0$} \\
        GCN      & $85.0 \pm 0.4$  & $75.3 \pm 0.5$  & $69.6 \pm 0.5$  & $60.9 \pm 0.7$  & $54.2 \pm 0.6$ & $48.4\pm0.5$ \\
        APPNP    & $\mathbf{86.3 \pm 0.4}$  & $75.8 \pm 0.5$  & $69.7 \pm 0.7$  & $60.3 \pm 0.9$  & $53.8 \pm 1.2$ & $49.0\pm1.6$\\
        GAT  & $83.5 \pm 0.5$  & $75.8 \pm 0.8$  & $71.2 \pm 1.2$  & $65.0 \pm 0.9$  & $60.5 \pm 0.9$  & $56.7 \pm 0.9$ \\
        \midrule
        GNNGuard & $83.1 \pm 0.7$  & $74.6 \pm 0.7$  & $70.2 \pm 1.0$  & $63.1 \pm 1.1$  & $57.5 \pm 1.6$ & $51.0\pm1.2$\\
        RGCN     & $85.7 \pm 0.4$  & $75.0 \pm 0.8$  & $69.1 \pm 0.4$  & $59.8 \pm 0.7$  & $52.8 \pm 0.7$ & $46.1\pm0.7$ \\
        GRAND    & \underline{$86.1 \pm 0.7$}  & $76.2 \pm 0.8$  & $70.7 \pm 0.7$  & $61.6 \pm 0.7$  & $56.7 \pm 0.8$ & $51.9\pm0.9$ \\
        ProGNN  & $85.6 \pm 0.5$  & $76.5 \pm 0.7$  & $71.0 \pm 0.5$  & $63.0 \pm 0.7$  & $56.8 \pm 0.7$  & $51.3 \pm 0.6$ \\
        Jaccard-GCN  & $83.7 \pm 0.7$  & $73.9 \pm 0.5$  & $68.3 \pm 0.7$  & $60.0 \pm 1.1$  & $54.0 \pm 1.7$  & $49.1 \pm 2.4$ \\

        GARNET  & $84.0 \pm 0.5$  & $76.5 \pm 0.4$  & $72.1 \pm 0.1$  & $66.4 \pm 0.7$  & $62.1 \pm 1.3$  & $58.7 \pm 1.5$ \\

        HANG &$84.5 \pm 0.2$&$75.7 \pm 0.6$ & $74.2 \pm 0.3$ & $69.5 \pm 0.4$ & $64.7 \pm 0.9$ & $58.4 \pm 0.1$\\
        NoisyGNN &$83.9\pm 0.5$ &$76.7\pm 0.1$&$72.1\pm 0.3$&$64.7\pm 0.7$&$58.0\pm 0.2$&$53.3\pm 0.6$\\
        EvenNet & $84.8 \pm 0.9$  & $75.8 \pm 0.9$  & $70.7 \pm 0.6$  & $63.9 \pm 0.5$  & $58.7 \pm 0.7$  & $54.7 \pm 0.8$ \\
        GraphCON &$83.7\pm 0.6$ & $ 75.8 \pm 0.3$ & $ 70.9 \pm 0.8$ & $ 65.9 \pm 0.7$ & $ 62.2 \pm 0.9$ & $ 56.6 \pm 0.3$\\
        SoftMedian  & $85.0 \pm 0.7$  & $78.6 \pm 0.3$  & $\underline{75.5 \pm 0.9}$  & \underline{$69.5 \pm 0.5$}  & $62.8 \pm 0.8$ & $58.1\pm0.7$ \\ 
        TWIRLS & $84.2 \pm 0.6$  & $77.3 \pm 0.8$  & $72.9 \pm 0.3$  & $66.9 \pm 0.2$  & $62.4 \pm 0.6$ & $58.7\pm1.1$\\
        TWIRLS-T  & $82.8 \pm 0.5$  & $76.8 \pm 0.6$  & $73.2 \pm 0.4$  & $67.7 \pm 0.4$  & $63.8 \pm 0.2$  & $60.8 \pm 0.3$ \\ 
        \midrule
        RUNG-$\ell_{1}$ (Ours) & $85.8 \pm 0.5$  & $\underline{78.4 \pm 0.4}$  & $74.3 \pm 0.3$  & $68.1 \pm 0.6$  & $63.5 \pm 0.7$ & $59.8\pm0.8$\\ 
        RUNG (Ours) & $84.6 \pm 0.5$  & $\mathbf{78.9 \pm 0.4}$  & $\mathbf{75.7 \pm 0.2}$  & $\mathbf{71.8 \pm 0.4}$  & $\mathbf{67.8 \pm 1.3}$ & $\mathbf{65.1 \pm 1.2}$\\ 
        \bottomrule
    \end{tabular}
    }
    \label{cora_adaptive_global_evasion}
\end{table}

Here we evaluate the the performance of RUNG against the baselines under different settings. The results of local and global adaptive attacks on Cora ML are presented in \autoref{cora_adaptive_local_evasion} and \autoref{cora_adaptive_global_evasion}, while those on Citeseer are presented in \autoref{citeseer_adaptive_local_evasion} and \autoref{citeseer_adaptive_global_evasion} in \autoref{sec:appendix_additional_results} due to space limits. 
We summarize the following analysis from Cora ML,
noting that the same observations apply to Citeseer.

\begin{itemize}[leftmargin=*,itemsep=0pt]

\item Under adaptive attacks, many existing defenses are not significantly more robust than undefended models. 
The $\ell_1$-based models such as TWIRLS, SoftMedian, and RUNG-$\ell_1$ demonstrate considerable and closely aligned robustness under both local and global attacks, which supports our unified $\ell_1$-based robust view analysis in Section~\ref{sec:unified-view}.

\item RUNG exhibits significant improvements over all baselines across various budgets under both global and local attacks. 
Local attacks are stronger than global attacks since local attacks concentrate on targeted nodes. The robustness improvement of RUNG appears to be more remarkable in local attacks.

\item When there is no attack, RUNG largely preserves an excellent clean performance. RUNG also achieves state-of-the-art performance under small attack budgets.

\end{itemize}

\subsection{Ablation study}
\label{sec:ablation}

\begin{figure}[t!] 
    \centering 
    \begin{minipage}[t]{0.32\textwidth}
        \centering
        \vspace{15pt}
        \includegraphics[width=0.9\textwidth]{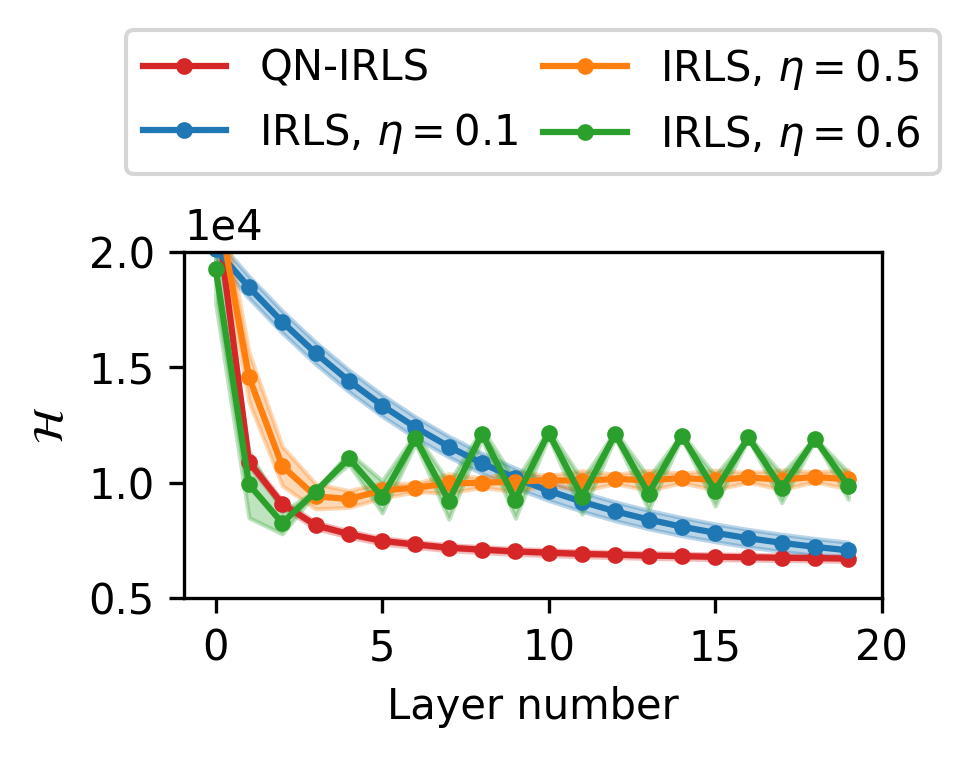}
        \vspace{-1.5ex}
        \caption{\centering Convergence of our QN-IRLS compared to IRLS. }
        \label{fig:convergence_of_ours}
    \end{minipage}
    \hspace{0pt}
\begin{minipage}[t]{0.32\textwidth} 
    \centering
    \vspace{20.13pt}
\includegraphics[width=0.923\textwidth]{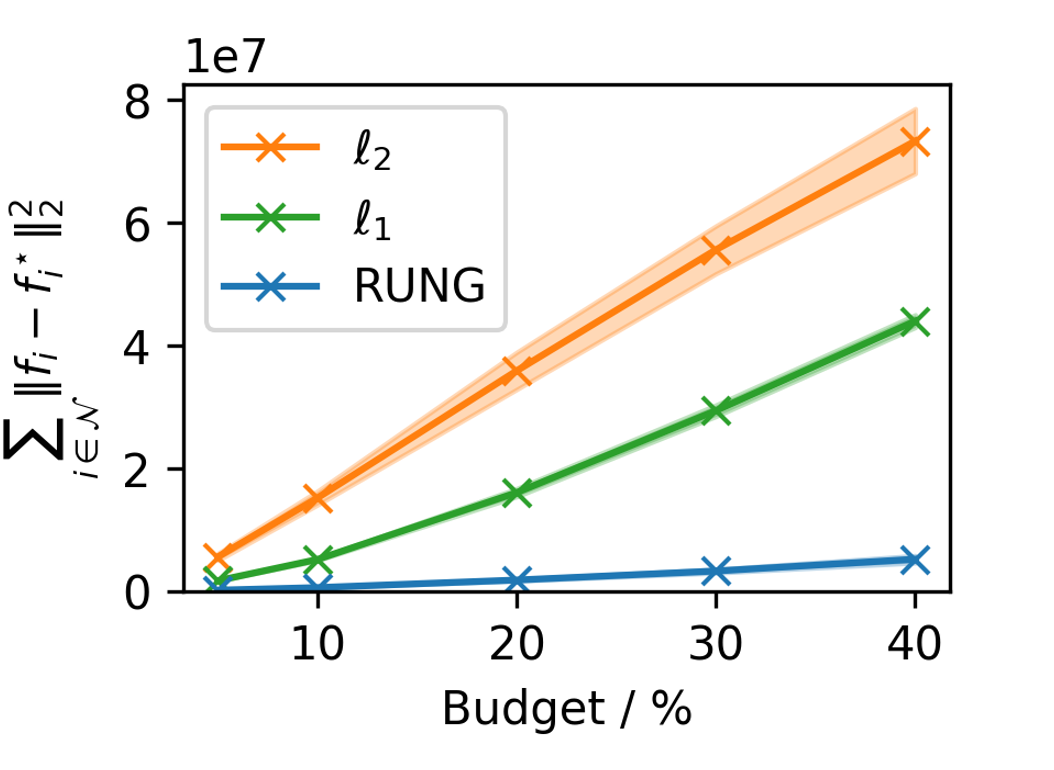}
    \vspace{-1.5ex}
    \caption{\centering Bias induced by \qquad different attack budgets. }
\label{fig:estimation_bias}
\end{minipage} 
    \hspace{0pt} 
    \begin{minipage}[t]{0.32\textwidth} 
        \centering
        \vspace{25.9pt}
        \includegraphics[width=0.9\textwidth]{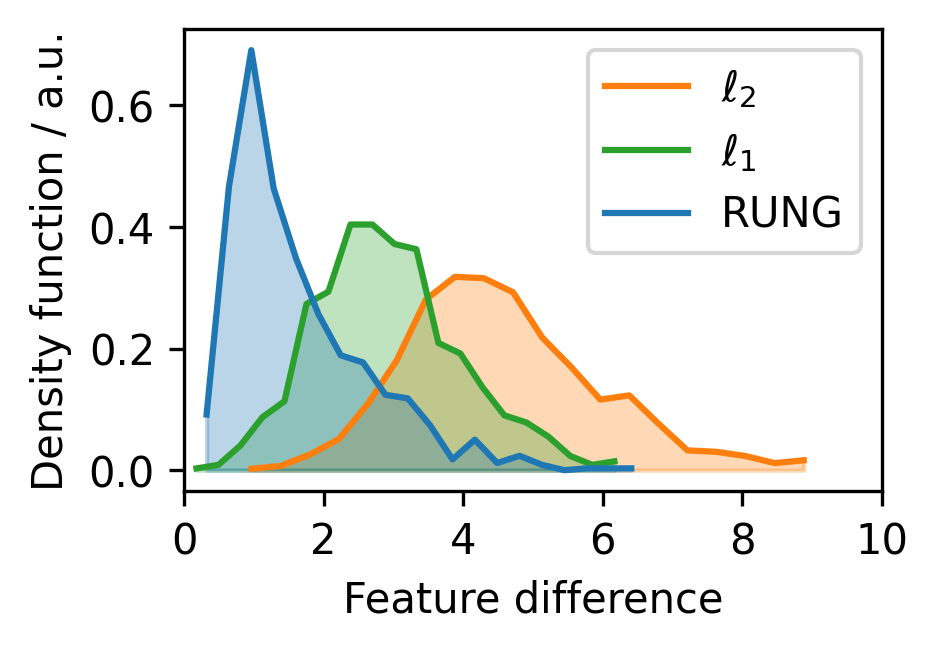}
               \vspace{-1.5ex}
        \caption{\centering Distribution of feature  difference on attacked edges.}
\label{fig:atk_edge_distribution}
    \end{minipage}
    % \vspace{-2pt}
\end{figure}

\textbf{Convergence.}
To verify the advantage of our QN-IRLS method in Eq~\eqref{eq:rw_update_f2} over the first-order IRLS in Eq~\eqref{eq:rw_update_f1}, we show the objective $\energy$ on each layer in \autoref{fig:convergence_of_ours}. It can be observed that our stepsize-free QN-IRLS 
% captures the landscape well, 
demonstrates the best convergence as discussed in Section~\ref{sec:algo}.

\textbf{Estimation bias.}
The bias effect in $\ell_1$-based GNNs and the unbiasedness of our RUNG can be empirically verified.
We quantify the bias with $\sumnode \|\vf_i-\vf_i^\star\|_2^2$, 
where $\vf_i^\star$ and $\vf_i$ denote the aggregated feature on the clean graph and attacked graph, respectively. 
As shown in \autoref{fig:estimation_bias}, when the budget scales up, $\ell_1$ GNNs exhibit a notable bias, whereas RUNG has nearly zero bias. We provide comprehensive discussion of unbiasedness of RUNG in Appendix~\ref{sec:bias_analysis_compre}.

\textbf{Defense Mechanism} 
To further investigate how our defense takes effect, 
we analyze the edges added under adaptive attacks. The distribution of the node feature differences $\big\|\vf_i/{\sqrt{d_i}} - \vf_j/\sqrt{d_j}\big\|_2$ of attacked edges 
is shown in \autoref{fig:atk_edge_distribution} for different graph signal estimators. 
It can be observed that our RUNG forces the attacker to focus on the edges with a small feature difference, indicating that our RUNG can improve robustness by down-weighting or pruning some malicious edges that connect distinct nodes.
Therefore, the attacks become less influential, which explains why RUNG demonstrates outstanding robustness.

\textbf{Transfer Attacks.} 
In addition to the adaptive attack, we also conduct a set of transfer attacks that take every baseline GNN as the surrogate model to comprehensively test the robustness of RUNG, following the unit test attack protocol proposed in \citep{mujkanovic2022_are_defenses_for_gnns_robust}. 
We summarize the results on Cora ML and Citeseer in~\autoref{fig:cora_transfer_global_evasion_to_mcp}  and ~\autoref{fig:citeseer_transfer_global_evasion_to_mcp} in \autoref{sec:appendix_additional_results} due to space limits. 
All transfer attacks are weaker than the adaptive attack in Section~\ref{sec:exp_main},  indicating the necessity to evaluate the strongest adaptive attack to avoid the false sense of security emphasized in this paper.
Note that the attack transferred from RUNG model
is slightly weaker than the adaptive attack since the surrogate and victim RUNG models have different model parameters in the transfer attack setting.

\textbf{Hyperparameters.} Due to the space limit, we provide the additional ablation studies on the hyperparameters (including $\gamma$ and $\lambda$ in MCP as well as the number of layers) of RUNG in Appendix~\ref{sec:additonal_ablation}. The results offer an overview strategy for the choice of optimal hyperparameters. We can observe that $\gamma$ in MCP has a significant impact on the performance of RUNG. Specifically, a larger $\gamma$ makes RUNG closer to an $\ell_1$-based model, while a smaller $\gamma$ encourages more edges to be pruned. This pruning helps RUNG to remove more malicious edges and improve robustness, although a small $\gamma$ may slightly reduce clean performance.

\textbf{Robustness under various scenarios.} Besides the evaluation under strong adaptive attacks, we also validate the consistent effectiveness of our proposed RUNG under various scenarios, including \textit{Transfer attacks} (Appendix~\ref{sec:transfer_attacks}), \textit{Poisoning attacks} (Appendix~\ref{sec:poisoning_attack}), \textit{Large scale datasets} (Appendix~\ref{sec:arxiv}), \textit{Adversarial training} (Appendix~\ref{sec:adv_train}), \textit{Graph injection attacks} (Appendix~\ref{sec:graph_injection_attack}).

% \vspace{-0.1in}
\section{Related Work}
% \vspace{-0.1in}
%\vspace{-0.75ex}
To the best of our knowledge, although there are works unifying existing GNNs from a graph signal denoising perspective~\citep{a_unifed_view_on_GNN_as_graph_denoising}, no work has been dedicated to uniformly understand the robustness and limitations of robust GNNs such as SoftMedian~\citep{softmedian}, SoftMedoid~\citep{softmedoid}, TWIRLS~\citep{TWIRLS}, ElasticGNN~\citep{elastic_gnn}, and TVGNN~\citep{hansen2023tvgnn}
% as an $\ell_1$ graph estimator 
from the $\ell_1$ robust statistics and bias analysis perspectives. 
To mitigate the estimation bias, MCP penalty is promising since it is well known for its near unbiasedness property \citep{mcp_regression} and has been %. %MCP has been  
applied to the graph trend filtering problem~\citep{varma2019vector} to promote piecewise signal modeling, but their robustness is unexplored. 
Nevertheless, other robust GNNs have utilized alternative penalties that might alleviate the bias effect. For example, GNNGuard~\citep{zhang2020gnnguard} prunes the edges whose cosine similarity is too small.
% , but its heuristic design is still vulnerable to adaptive attacks. 
Another example is that TWIRLS~\citep{TWIRLS} with a thresholding penalty can also exclude edges using graph attention.
However, the design of their edge weighting or graph attention is heuristic-based and exhibits suboptimal performance compared to the RUNG proposed in this work. 

\section{Conclusion \& Limitation}
\label{sec:conclusion}
In this work, we propose
a unified view of $\ell_1$ robust graph smoothing to uniformly understand the robustness and limitations of
representative robust GNNs.
The established view not only justifies their improved and closely aligned robustness but also explains their severe performance degradation under large attack budgets by 
a novel estimation bias analysis.
To mitigate the estimation bias, we propose a robust and unbiased graph signal estimator. 
To solve this non-trivial estimation problem, we design a novel and efficient Quasi-Newton IRLS algorithm 
that can better capture the landscape of the optimization problem and converge stably with a theoretical guarantee. 
This algorithm can be unfolded and used as a building block for constructing robust GNNs with Robust Unbiased Aggregation (RUNG).
As verified by our experiments, RUNG provides the best performance under strong adaptive attacks among all the baselines. 
Furthermore, RUNG also covers many classic GNNs as special cases.
Most importantly, this work provides a deeper understanding of existing approaches 
and reveals a principled direction for designing robust GNNs. 

Regarding the limitations, first, the improvement of RUNG is more significant under large budgets compared to the robust baselines. Second, we primarily include experiments on homophilic graphs in the main paper, but we can generalize the proposed robust aggregation to heterophilic graphs in future work. Third, although our Quasi-Newton IRLS algorithm has exhibited excellent convergence compared to the vanilla IRLS, the efficiency of RUNG could be further improved.

\section*{Acknowledgment}
Zhichao Hou and Dr. Xiaorui Liu are supported by the National Science Foundation (NSF) National AI Research Resource Pilot Award, Amazon Research Award, NCSU Data Science Academy Seed Grant Award, and NCSU Faculty Research and Professional Development Award.

% \newpage
\bibliography{ref}
\bibliographystyle{ref}

\appendix

% formal

% \section{Iterative Reweighted Least Square}\label{sec:convergence}
\newpage
\appendix
\onecolumn

\section{Bias Accumulation of L1 Models}
\label{sec:bias_details}

\subsection{Details of the Numerical Simulation Settings}

To provide a more intuitive illustration of 
the estimation biases of 
%$\ell_1$-based models, 
different models, we simulate a mean estimation problem on synthetic data since most message passing schemes in GNNs essentially estimate the mean of neighboring node features.
In the context of mean estimation, the bias is measured as the distances between different mean estimators and the true mean. % I think this seems obvious
We firstly generated clean samples $\{\vx_i\}_{i=1}^{n}$ (blue dots) and the outlier samples $\{\vx_i\}_{i=n+1}^{n+m}$(red dots) from 2-dimensional Gaussian distributions, $\cN((0, 0), 1)$ and $\cN((8, 8), 0.5)$, respectively. We calculate the mean of clean samples $\smash[b]{\frac{1}{n}\sum_{i=1}^n\vx_i}$ as the ground truth of the mean estimator. Then we estimate the mean of all the samples by solving
$\argmin_\vz \sum_{i=1}^{n+m} \eta(\vz - \vx_i)$ using the Weiszfeld method
% which is solved by reweighting $\vz$
\citep{enhancing_l1_with_reweighting,weiszfeld_old_and_new}, where $\eta(\cdot)$ can take different norms such as $\ell_2$ norm $\|\cdot\|_2^2$ and $\ell_1$ norm $\|\cdot\|_2$. 
%The detailed simulation settings and results are available in Appendix~\ref{sec:bias_details}.
% $\|\cdot\|_2^2$, $\|\cdot\|_1$ or $\|\cdot\|_2$.
%%%
% \ruiqi{solution using Weiszfield, cite}

%In \autoref{sec:bias}, we conducted a numerical simulation of mean estimation on synthetic data $\vx_i$. 
The mean estimators are formulated as minimization operators
\begin{equation}
    \bar\vz = \argmin_{\vz} \sum_{i}^{n+m} \eta(\vz - \vx_i),
\end{equation}
where $n$ is the number of clean samples and $m$ is the number of adversarial samples. 

\textbf{$\ell_1$ estimator.} The $\ell_1$ estimator ($\eta(\vy) \coloneqq \|\vy\|_2$), essentially is the geometric median. We adopted the Weiszfeld method to iteratively reweight $\vz$ to minimize the objective, following
\begin{equation}\label{eq:weiszfeld_mest}
    \vz^{(k+1)} =  \frac{\sum_i w_i^{(k)}\vx_i}{\sum_i w_i^{(k)}},
\end{equation}
where $w_i^{(k)}=\frac{1}{\|\vz^{(k)} - \vx_i\|_2}$. This can be seen as a gradient descent step of $\vz^{(k+1)} = \vz^{(k)} - \alpha\nabla_\vz \sum_i\|\vz - \vx_i\|_2 = \vz^{(k+1)} - \alpha \sum_i \frac{\vz^{(k)} - \vx_i}{\|\vz^{(k)} - \vx_i\|_2}$. Taking $\alpha=\frac{1}{\sum_i w_i^{(k)}}$ instantly yields Eq. \eqref{eq:weiszfeld_mest}.

\textbf{MCP-based estimator.} We therefore adopt a similar approach for the MCP-based estimator (``Ours'' in Fig. \autoref{fig:m_estimator_compare}), where $\eta(\vy) \coloneqq \rho_\gamma(\vy)$:
\begin{align}
    \vz^{(k+1)} &= \vz^{(k)} - \alpha\nabla_\vz \sum_i\rho_\gamma(\|\vz - \vx_i\|_2)\\
    &= \vz^{(k)} - \alpha \sum_i \max(0, \frac{1}{\|\vz^{(k)} - \vx_i\|_2} - \frac{1}{\gamma}) (\vz^{(k)} - \vx_i).
\end{align}
Denoting $\max(0, \|\vz^{(k)} - \vx_i\|_2^{-1} - \frac{1}{\gamma})$ as $w_i$, and then $\alpha =  \frac{1}{\sum_i w_i}$ yields a similar reweighting iteration $\vz^{(k+1)} =  \frac{\sum_i w_i^{(k)}\vx_i}{\sum_i w_i^{(k)}}$.

\paragraph{$\ell_2$ estimator.} It is worth noting that the same technique can be applied to the $\ell_2$ estimator with $\rho(\vz)\coloneqq\|\vz\|_2^2$. The iteration becomes
\begin{align}
        \vz^{(k+1)} &= \vz^{(k)} - \alpha\nabla_\vz \sum_i\|\vz - \vx_i\|_2^2\\
    &= \vz^{(k)} - \alpha \sum_i (\vz^{(k)} - \vx_i),
\end{align}
and $\alpha=\frac{1}{n+m}$ yields $\vz^{(k+1)} = \frac{1}{n+m}\sum_i \vx_i$, which gives the mean of all samples in one single iteration.

Similarities between this mean estimation scenario and our QN-IRLS in graph smoothing can be observed, both of which involve iterative reweighting to estimate similar objectives. The approximated Hessian in our QN-IRLS resembles the Weiszfeld method, canceling the $\vz^{(k)}$ by tuning the stepsize.

In Figure~\ref{fig:m_estimator_compare}, we visualize the generated clean samples and outliers, as well as the ground truth means and the mean estimators with $\eta(\cdot)=\|\cdot\|_2^2$ or $\|\cdot\|_2$ under different outlier ratios (15\%, 30\%, 45\%). 
% As shown in \autoref{fig:m_estimator_compare}, 
The results show that the $\ell_2$-based estimator deviates far from the true mean, while the $\ell_1$-based estimator is more resistant to outliers, which explains why $\ell_1$-based methods exhibit stronger robustness.
However, as the ratio of outliers escalates, the $\ell_1$-based estimator encounters a greater shift from the true mean due to the accumulated bias caused by outliers.
This observation explains why $\ell_1$-based graph smoothing models suffer from catastrophic degradation under large attack budgets. Our estimator keeps much closer to the ground truth than other estimators with the existence of outliers.

\subsection{Additional Simulation Results and Discussions}
Here, we complement \autoref{fig:m_estimator_compare} with the settings of higher attack budgets. As the outlier ratio exceeds the breakdown point $50\%$, we observe that our MCP-based mean estimator can correctly recover the majority of the samples, i.e. converge to the center of ``outliers''.
\begin{figure}[h!]
    \centering
\includegraphics[width=1.0\textwidth]{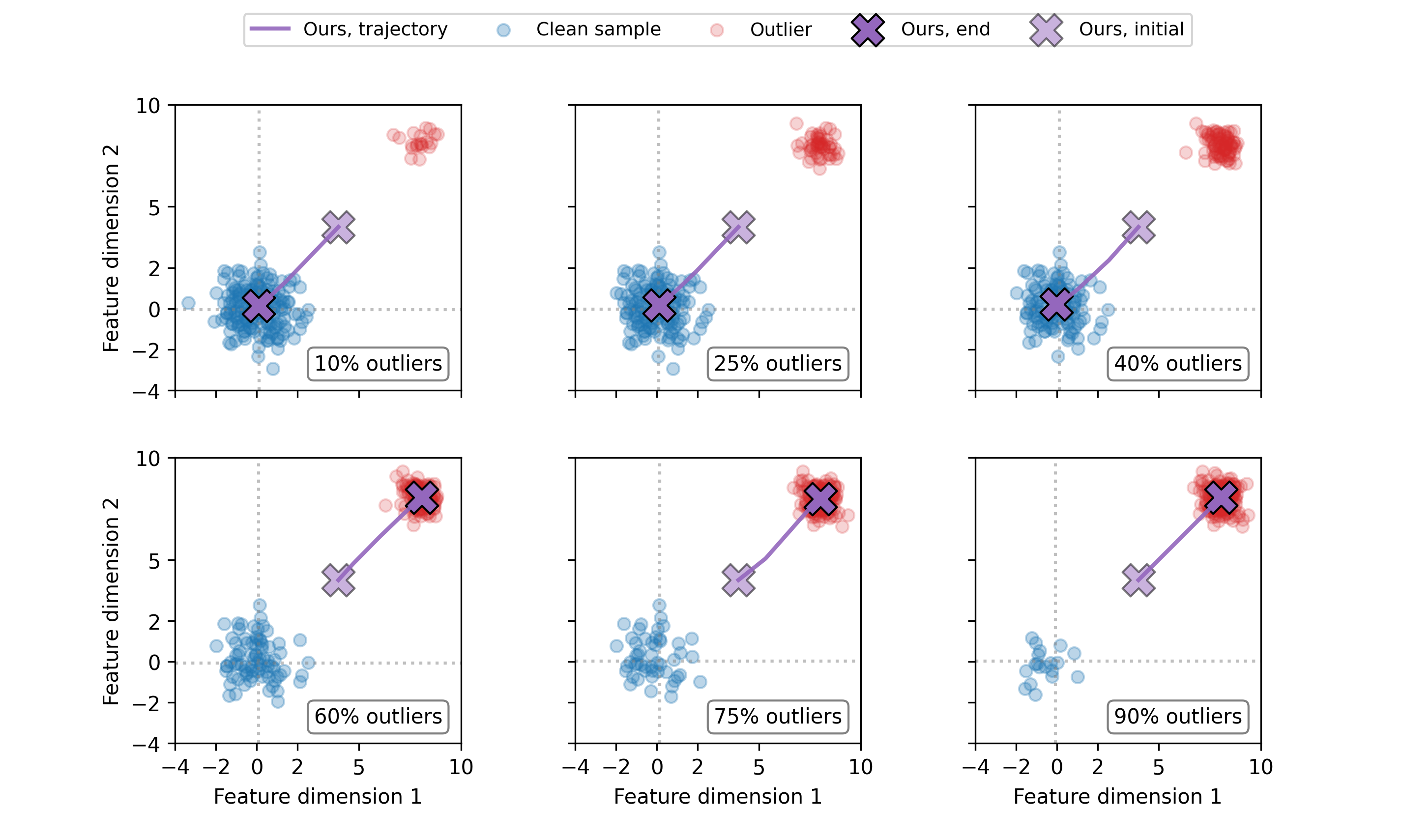}
    % \vspace{-0.1in}
    \caption{The trajectory of our MCP-based mean estimator.}
    \label{fig:mest_appendix_init_4_traj_global}
\end{figure}

\section{Convergence Analysis}\label{sec:convergence}

To begin with, 
we will provide an overview of our proof, followed by a detailed presentation of the formal proof for the convergence analysis.

\textbf{Overview of proof.}
First, for both IRLS and QN-IRLS, 
we construct, for $\mF^{(k)}$ in every iteration $k$, a quadratic upper bound $\Hhat$ that satisfies $\Hhat + C\ge\energy$ where the equality is reached at $\mF^{(k)}$. 
%Borrowing the spirit from the MM-algorithm, 
Then we can minimize $\Hhat$ to guarantee the iterative descent of $\energy$ since $\energy(\mF^{(k+1)}) \le \Hhat(\mF^{(k+1)}) + C \le \Hhat(\mF^{(k)}) + C=\energy(\mF^{(k)})$.

% To minimize $\Hhat$, we utilize a diagonal matrix $\hat{\mQ}$ to define the update iteration as $\mF^{(k+1)}= \mF^{(k)} - \hat{\mQ}^{-1}\nabla\Hhat(\mF^{(k)})$, which can guarantee an efficient descent of $\Hhat$.
To find the $\mF^{(k+1)}$ such that $\Hhat(\mF^{(k+1)}) \le \Hhat(\mF^{(k)})$, IRLS simply adopts the plain gradient descent $\mF^{(k+1)}= \mF^{(k)} - \eta\nabla\Hhat(\mF^{(k)})$ whose convergence condition can be analyzed with the $\beta$-smoothness of the quadratic $\Hhat$ (\autoref{theorem:gd_descent}). To address the problems of IRLS as motivated in Section~\ref{sec:qn-irls}, our Quasi-Newton IRLS utilizes the diagonal approximate Hessian $\hat{\mQ}$ to scale the update step size in different dimensions respectively as $\mF^{(k+1)}= \mF^{(k)} - \hat{\mQ}^{-1}\nabla\Hhat(\mF^{(k)})$. Thereafter, by bounding the Hessian with $2\hat\mQ$, the descent condition of $\Hhat$ is simplified (\autoref{theorem:preconditioned_descent}).
% \ruiqi{but there are two theorems, shall we mention that both theorem 1 and 2 follow from the same mm-algo's ideology?}

%Lemma~\ref{lemma:weighted_majorizer}: 

%To minimize $\Hhat$, we construct an even simpler upper bound of $\Hhat$ by finding a diagonal matrix $\mathcal{Q}$ such that $\mathcal{Q}\succeq\nabla^2\Hhat$ using basic properties of the graph. This naturally motivates us to use $\hat\mQ \propto \mathcal{Q}$. 
% The coefficient is coupled with $\eta$ and is taken as $\frac{1}{4}$ in our formulation.
% \xr{tell some techniques in high-level to show the novelty in analysis}

\begin{lemma}\label{lemma:weighted_majorizer}
        %For any node features $\mF$ and 
        For any $\rho(y)$ satisfying $\frac{d\rho(y)}{dy^2}$ is non-increasing, denote $y_{ij}:=\left\|\frac{\vf_i}{\sqrt{d_i}} - \frac{\vf_j}{\sqrt{d_j}}\right\|_2$, then
        $\energy(\mF) = \sumedgehollow\rho(y_{ij}) + \lambda\sumnode \|\vf_i - \vf_i^{(0)}\|_2^2$
         has the following upper bound:
        \begin{equation}\label{eq:converge_mm_obj}
            \energy(\mF) \le \Hhat(\mF) + C = \sumedgehollow \mW_{ij}^{(k)}
            % \|\frac{\vf_i}{\sqrt{d_i}} - \frac{\vf_j}{\sqrt{d_j}}\|_2^2
        y_{ij}^2 + \lambda\sumnode \|\vf_i - \vf_i^{(0)}\|_2^2 + C, 
        \end{equation}
        where $\mW_{ij}^{(k)} = \frac{\partial \rho(y)}{\partial y^2}\big{|}_{y = y_{ij}^{(k)}
        % \|\vf_i^{(k)}/\sqrt{d_i} - \vf_j^{(k)}/\sqrt{d_j}\|_2}
        }$ and 
        $y_{ij}^{(k)} = \left\|\frac{\vf_i^{(k)}}{\sqrt{d_i}} - \frac{\vf_j^{(k)}}{\sqrt{d_j}}\right\|_2$ 
        and $C = \energy(\mF^{(k)}) - \Hhat(\mF^{(k)})$ is a constant.  The equality in Eq.~\eqref{eq:converge_mm_obj} is achieved when $\mF = \mF^{(k)}$. %For simplicity, we denote $\Hhat^{(k)}$  as $\Hhat$ .
    \end{lemma}
    \begin{proof}
        % Now we prove that Eq.~\eqref{eq:converge_mm_obj} holds. A sufficient condition is that
        Let $v=y^2$ and define $\psi(v):=\rho(y)=\rho(\sqrt{v})$. Then $\psi$ is concave since $\frac{d\psi(v)}{dv}=\frac{d\rho(y)}{dy^2}$ is non-increasing. According to the concavity property, we have $\psi(v) \le \psi(v_0) + \psi'(\nu)\big{|}_{\nu=v_0} (v-v_0)$. 
        Substitute $v=y^2,v_0=y_0^2$, we obtain:
        \begin{align}\label{eq:lemma_mm_onedim}
             \rho(y) &\le y^2 {\frac{\partial \rho(y)}{\partial y^2}\big|_{y=y_0}} - y_0^2 \frac{\partial \rho(y)}{\partial y^2}\big|_{y=y_0}+ {\rho(y_0)}\\
             &= y^2 {\frac{\partial \rho(y)}{\partial y^2}\big|_{y=y_0}}+C(y_0)
        \end{align}
        where the inequality is reached when $y = y_0$.
        Next, substitute $y=y_{ij}$ and $y_0=y_{ij}^{(k)}$,  we can get $\rho(y_{ij})\le\mW_{ij}^{(k)}y_{ij}^2+C(y_{ij}^{(k)})$ which takes the equality at $y_{ij}=y_{ij}^{(k)}$. Finally, by summing up both sides and add a regularization term, we can prove the Eq.~\eqref{eq:converge_mm_obj}.
    \end{proof}
    \begin{remark}
        It can be seen that the definition of $\Hhat$ depends on $\mF^{(k)}$, which ensures that the bound is tight when $\mF=\mF^{(k)}$. This tight bound condition is essential in the majorization-minimization algorithm as seen  in \autoref{theorem:gd_descent}.
    \end{remark}

    \begin{lemma}\label{lemma:gradien_and_hessian_of_Hhat}
        For $\Hhat = \sumedgehollow \mW_{ij} y_{ij}^2 + \lambda\sumnode \|\vf_i - \vf_i^{(0)}\|_2^2$, the gradient and Hessian w.r.t. $\mF$\footnote{Here are some explanations on the tensor `Hessian' $\nabla^2\Hhat$. Since $\Hhat(\mF)$ is dependent on a matrix, there are some difficulties in defining the Hessian. However, as can be observed in Eq.~\eqref{eq:rw_appnp_energy_grad_hollow_diag} and Eq.~\eqref{eq:rw_appnp_energy_hessian_tensor}, the feature dimension can be accounted for by the following. Initially, we treat the feature dimension as an irrelevant dimension that is excluded from the matrix operations. E.g., $\mF\nabla^2\Hhat\mF = \sum_{ik}\mF_{ij}\nabla^2\Hhat_{ijkl}\mF_{kl}$ where the feature dimensions $j$ and $l$ remain free indices while the node indices $i$ and $k$ are eliminated as dummy indices. Finally, we take the trace of the resulting \#feature$\times$\#feature matrix to get the desired value.\label{footnote:explain_hessian_tensor}} satisfy
        \begin{equation}
            \nabla_{\mF_{mn}} \Hhat(\mF) = 2\left( (\text{diag}(\vq) - \mW\odot\tilde{\mA} + \lambda \mI)  \mF  - \lambda\mF^{(0)}\right)_{mn},
        \end{equation}
        and \begin{equation}
            \nabla_{\mF_{mn}}\nabla_{\mF_{kl}} \Hhat(\mF) = 2\left ( \text{diag}(\vq) - \mW\odot\tilde{\mA} + \lambda\mI\right)_{mk},
        \end{equation}
        where $\vq_m = \sum_{j} \mW_{mj}\mA_{mj} / d_m$ and $\tilde{\mA}_{ij} = \frac{\mA_{ij}}{\sqrt{d_id_j}}$ is the symmetrically normalized adjacency matrix.
        
    \end{lemma}

    \begin{proof}
         Follow $\mA = \mA^\top$ and define $~y^2_{ij}\coloneqq\left\|\frac{\vf_i}{\sqrt{d_i}} - \frac{\vf_j}{\sqrt{d_j}}\right\|^2_2 $, then the first-order gradient of $\sumedgehollow \mW_{ij}y_{ij}^2$ will be 
        \begin{align}
            &\nabla_{\mF_{mn}} \left(\sumedgehollow \mW_{ij}y_{ij}^2\right) \\ 
            \label{eq:W_hollow_empty_sum}= & \sumedgehollow \mW_{ij}\frac{\partial y^2_{ij}}{\partial \mF_{mn}} \\
            =&\sum_{(m,j)\in\mathcal{E}} \mW_{mj}\frac{\partial y^2_{mj}}{\partial \mF_{mn}}\\
            =&\sum_{(m,j)\in\mathcal{E}} \mW_{mj}\frac{\partial \left(\frac{\mF_{mn}}{\sqrt{d_m}} - \frac{\mF_{jn}}{\sqrt{d_j}}\right)^2}{\partial \mF_{mn}}\\
            \label{eq:W_hollow_empty_W}= & \sum_{j\in \mathcal{N}(m)} 2\mW_{mj} (\frac{\mF_{mn}}{d_m}  - \frac{\mF_{jn}}{\sqrt{d_md_j}} ) \\
            = & 2 \sum_{j} \mW_{mj} (\frac{\mA_{mj}}{d_m} \mF_{mn} - \frac{\mA_{mj}}{\sqrt{d_md_j}}\mF_{jn}) \\
            = & 2 (\frac{\sum_{j} \mW_{mj}\mA_{mj}}{d_m}) \mF_{mn} - 2((\mW\odot\tilde{\mA}) \mF)_{mn} \\
            = & \left (2 (\text{diag}(\vq) - \mW\odot\tilde{\mA}) \mF \right)_{mn}
            \label{eq:rw_appnp_energy_grad_hollow_diag},
        \end{align}
        and the second-order hessian will be:
        \begin{align}
            &\nabla^2_{\mF_{mn}\mF_{kl}} \left(\sumedgehollow \mW_{ij}y_{ij}^2\right)\\
            = & \sumedgehollow \mW_{ij}\frac{\partial y^2_{ij}}{\partial \mF_{mn} \partial \mF_{kl}} \\
            = & 2\frac{\partial}{\partial \mF_{kl}} \left(\sum_{j} \mW_{mj} (\frac{\mA_{mj}}{d_m} \mF_{mn} - \frac{\mA_{mj}}{\sqrt{d_md_j}}\mF_{jn})\right) \\
            = & 2 (\vq_m\delta_{mk} - \sum_j \mW_{mj}\frac{\mA_{mj}}{\sqrt{d_md_j}}\delta_{jk} )\delta_{nl} \\ 
            = & 2 (\text{diag}(\vq) -  \mW\odot\tilde{\mA})_{mk}\delta_{nl} .\label{eq:rw_appnp_energy_hessian_tensor}
        \end{align}
    \end{proof}
    
    \begin{remark}\label{remark:hollow_W}
        From Eq.~\eqref{eq:W_hollow_empty_sum} to Eq.~\eqref{eq:W_hollow_empty_W}, one can assume $m\notin\mathcal{N}(m)$, and thus $\mW_{mm}=0$. However, as we know, a self-loop is often added to $\mA$ to facilitate stability by avoiding zero-degree nodes that cannot be normalized. This is not as problematic as it seems, though. Because $\sumedgehollow$ intrinsically excludes the diagonal terms, we can simply assign zero to the diagonal terms of $\mW$ so that the term of $j=m$ is still zero in Eq.~\eqref{eq:W_hollow_empty_W}, as defined in Eq.~\eqref{eq:Hhat}.

        %As an extra discussion, even if we change the objective from $\sumedgehollow (\cdot)$ to $\sumedge (\cdot)$ in Eq.~\eqref{eq:Hhat}, it is still an upper bound of $\energy$ and all the proof is the same except $\nabla\Hhat$ and $\nabla^2\Hhat$. The problem is that $\mW\coloneqq d_{y^2}\rho(y)$ is not defined at $y=0$. If we further define $d_{y^2}\rho(y)\big|_{y=0}\coloneqq\lim_{\xi\rightarrow 0^+}\frac{\rho(\xi)}{\xi^2}$, many penalties will explode such as our UGE or $\ell_1$ norm. If we again further take a threshold of $d_{y^2}\rho(y)\big|_{y=0}$, e.g. for $\ell_1$ estimator ($\rho(y)=y$), take $d_{y^2}\rho(y)\big|_{y=0}\coloneqq\max(\frac{\lambda}{\varepsilon},d_{y^2}\rho(y)\big|_{y=\varepsilon})$, then $\vq$ will be very large, which corresponds to a very small stepsize as will be seen in \autoref{theorem:preconditioned_descent}. Therefore, for the efficiency of our algorithm, we take the formulation of a zero-diagonal $\mW$.
    \end{remark}
    
    To minimize $\Hhat$, the gradient descent update rule takes the form of Eq.~\eqref{eq:rw_update_f1}.
    One may assume that when $\eta$ is chosen to be small enough, $\Hhat(\mF^{(k+1)})\le\Hhat(\mF^{(k)})$. 
    For a formal justification, we have \autoref{theorem:gd_descent} to determine the convergence condition of $\eta$.
    
    \begin{theorem}\label{theorem:gd_descent}
    If $\mF^{(k)}$ follows the update rule in Eq.~\eqref{eq:rw_update_f1}, where the $\rho$ satisfies that $\frac{d\rho(y)}{dy^2}$ is non-decreasing for $y\in(0,\infty)$, 
    then a sufficient condition for $\energy(\mF^{(k+1)})\le \energy(\mF^{(k)})$ is that the step size $\eta$ satisfies $0 < \eta \le \|\text{diag}(\vq^{(k)}) - \mW^{(k)}\odot\tilde{\mA} + \lambda\mI\|_2^{-1}.$
        % \begin{equation}
        %      0 < \eta \le \|\text{diag}(\vq) - \mW^{(k)}\odot\tilde{\mA} + \lambda\mI\|_2^{-1}.
        % \end{equation}    
    \end{theorem}
    \begin{proof}
        %As has been discussed in \autoref{sec:algo}, using \autoref{lemma:quadratic_hessian_upper_bound} 
        The descent of $\Hhat(\mF)$ can ensure the descent of $\energy(\mF)$ since $\energy(\mF^{(k+1)}) \le \Hhat(\mF^{(k+1)}) \le \Hhat(\mF^{(k)})=\energy(\mF^{(k)})$. Therefore, we only need to prove  $\Hhat(\mF^{(k+1)})\le\Hhat(\mF^{(k)})$.
        
        Noting that $\Hhat$ is a quadratic function and $\mF^{(k+1)} - \mF^{(k)} = -\eta\nabla \Hhat(\mF^{(k)})$, then $\Hhat(\mF^{(k+1)})$ and $\Hhat(\mF^{(k)})$ can be connected using Taylor expansion\footref{footnote:explain_hessian_tensor}, where $\nabla\Hhat$ and $\nabla^2\Hhat$ is given in \autoref{lemma:gradien_and_hessian_of_Hhat}:
    \begin{align}
        & \Hhat(\mF^{(k+1)}) - \Hhat(\mF^{(k)}) \\
        = & \tr \left(\nabla \Hhat(\mF^{(k)})^\top (\mF^{(k+1)} - \mF^{(k)})\right) \\
        & + \frac{1}{2}\tr \left( (\mF^{(k+1)} - \mF^{(k)})^\top \nabla^2 \Hhat(\mF^{(k)}) (\mF^{(k+1)} - \mF^{(k)}) \right)\\
        = &\tr\left(
            -\eta \nabla \Hhat(\mF^{(k)})^\top \nabla \Hhat(\mF^{(k)}) \right) \\
           & + \tr\left( \frac{\eta^2}{2}\nabla \Hhat(\mF^{(k)})^\top \nabla^2\Hhat(\mF^{(k)})\nabla \Hhat(\mF^{(k)})
        \right).
    \end{align}
    Insert $\nabla^2\Hhat(\mF^{(k)}) = 2(\text{diag}(\vq) - \mW^{(k)}\odot\tilde{\mA} + \lambda\mI)$ from \autoref{lemma:gradien_and_hessian_of_Hhat} into the above equation and we can find a sufficient condition for $\Hhat(\mF^{(k+1)}) \le \Hhat(\mF^{(k)})$ to be
    \begin{equation}
        -\eta + \|\text{diag}(\vq) - \mW^{(k)}\odot\tilde{\mA} + \hat{\lambda}\mI\|_2 \eta^2 \le 0,
    \end{equation}
    or 
    \begin{equation}
        \eta \le \|\text{diag}(\vq) - \mW^{(k)}\odot\tilde{\mA} + \hat{\lambda}\mI\|_2^{-1}.
    \end{equation}
    \end{proof}
    
    \par

    \iffalse
    So far, our algorithm can be summarized as follows. Considering the $k$-th iteration, where two steps take place:
    \begin{itemize}
        \item Majorization step Eq.~\eqref{eq:converge_mm_obj}. $\energy(\mF)$ is bounded by $\Hhat(\mF)$. They happen to equal at $\mF = \mF^{(k)}$
        \item Gradient descent step Eq.~\eqref{eq:rw_update_f1}. $\mF^{(k+1)}$ is found such that $\Hhat(\mF^{(k+1)})\le \Hhat(\mF^{(k)})$ and thus $\energy(\mF^{(k+1)})\le \energy(\mF^{(k)})$.
    \end{itemize}
    By iterating the above steps, $\energy$ can be minimized to a local minimum.
    \fi
    \par

    %However, the above convergence criterion is not satisfactory. When the input graph $\mA$ or the feature (which $\mW$ depends on) are different, the convergent step size is prone to constant changes. Moreover, for some data, the step size could even become too small to produce a converging solution in practical time. Therefore, we propose to use the following Quasi-Newton IRLS to address these problems.

    \par
    Now we prove that when taking the Quasi-Newton-IRLS step as in Eq.~\eqref{eq:rw_update_f2}, the objective $\Hhat$ is guaranteed to descend.
    %which is exactly \autoref{theorem:preconditioned_descent}.
    %Prior to that, we will be using \autoref{lemma:quadratic_hessian_upper_bound}. 
    Since the features in different dimensions are irrelevant, we simplify our notations as if feature dimension was $1$. 
    %We can then use vectors $\vx$ and $\vy$ %\footnote{Although with a slight abuse of notations: $y_{ij}$ is previously defined as the edge difference.} to replace the node feature matrix $\mF$, making the proof clearer without the loss of generality.
    One may 
    %refer to\footref{footnote:explain_hessian_tensor} to 
    easily recover the general scenario 
    %by simply adding extra feature dimensions to each vector and canceling these dimensions 
    by taking the trace.
    \begin{lemma}\label{lemma:quadratic_hessian_upper_bound}
    $2\hat{\mQ}- \nabla^2 \Hhat(\vy)$ is positive semi-definite, where $\Hhat=\sumedgehollow \mW_{ij} y_{ij}^2 + \lambda\sumnode \|\vf_i - \vf_i^{(0)}\|_2^2$, $\hat\mQ = 2(\text{diag}(\vq) + \lambda\mI)$, and $\vq_m = \sum_{j} \mW_{mj}\mA_{mj} / d_m$.
    \end{lemma}
    \begin{proof}
    In Lemma~\ref{lemma:gradien_and_hessian_of_Hhat}, we have $\nabla^2 \Hhat(\vy) = 2 (\text{diag}(\vq)  +\lambda \mI- \mW\odot\tilde{\mA})$, then
        \begin{equation}\label{eq:convergence_should_be_PSD}
             2\hat{\mQ} - \nabla^2 \Hhat(\vy) = 2 (\text{diag}(\vq) + \lambda\mI + \mW\odot\tilde{\mA}).
        \end{equation}
        %We are going to prove Eq.~\eqref{eq:convergence_should_be_PSD} is positive semi-definite by constructing a summation of quadratic terms. 
        Recall how we derived Eq.~\eqref{eq:rw_appnp_energy_grad_hollow_diag} from Eq.~\eqref{eq:converge_mm_obj}, where we proved that
        \begin{equation}
            \sumedgehollow \mW_{ij}\| \frac{\vf_i}{\sqrt{d_i}} - \frac{\vf_j}{\sqrt{d_j}} \|_2^2 = \tr(\mF^\top (\text{diag}(\vq) - \mW \odot \tilde{\mA}) \mF),  
        \end{equation}
        which holds for all $\mF$. Similarly, the equation still holds after flipping the sign before $\vf_j / \sqrt{d_j}$ and $\mW \odot \tilde{\mA}$. We then have this inequality: $\forall \mF, \forall\lambda \ge 0$
        \begin{align}
            0\le &\sumedgehollow \mW_{ij}\| \frac{\vf_i}{\sqrt{d_i}} + \frac{\vf_j}{\sqrt{d_j}} \|_2^2 = \tr(\mF^\top (\text{diag}(\vq) + \mW \odot \tilde{\mA}) \mF) \\
            \le & \tr(\mF^\top (\text{diag}(\vq) + \mW \odot \tilde{\mA} + \lambda\mI) \mF).
        \end{align}
        Thus, $(\text{diag}(\vq) + \mW \odot \tilde{\mA} + \lambda\mI) \succeq 0$, and thus $2\hat{\mQ} - \nabla^2 \Hhat(\vy)\succeq 0$.
        
    \end{proof}
    
    Using \autoref{lemma:weighted_majorizer} and \autoref{lemma:quadratic_hessian_upper_bound} we can prove \autoref{theorem:preconditioned_descent}. Note that we continue to assume \#feature$=1$ for simplicity but without loss of generality\footref{footnote:explain_hessian_tensor}.
    
    \begin{theorem}\label{theorem:preconditioned_descent}
        If $\mF^{(k+1)}$ follows update rule in Eq.~\eqref{eq:rw_update_f2}, where $\rho$ satisfies that $\frac{d\rho(y)}{dy^2}$ is non-decreasing $\forall y\in(0,\infty)$,
        it is guaranteed that $\energy(\mF^{(k+1)})\le \energy(\mF^{(k)})$. 
    \end{theorem}
    \begin{proof}
        Following the discussions in \autoref{theorem:gd_descent}, we only need to prove $\Hhat(\mF^{(k+1)})\le\Hhat(\mF^{(k)})$.% under the same conditions.
        For the quadratic $\Hhat$, we have:
        \begin{equation}\label{eq:prove_preconditioning_convergence_quadratic_taylor}
            \Hhat(\vx) = \Hhat(\vy) + \nabla \Hhat(\vy)^\top (\vx - \vy) + \frac{1}{2} (\vx - \vy)^\top \nabla^2 \Hhat(\vy) (\vx - \vy).
        \end{equation}
        We can define  $\mathcal{Q}(\vy)=2\hat{\mQ}(\vy)$ in \autoref{lemma:quadratic_hessian_upper_bound} such that $\mathcal{Q}(\vy) - \nabla^2 \Hhat(\vy) \succeq 0$, then
        \begin{equation}\label{eq:prove_preconditioning_convergence_hessian_bound}
            \forall \vz, \vz^\top\mathcal{Q}(\vy)\vz \ge \vz^\top\nabla^2 \Hhat(\vy)\vz.
        \end{equation}
        Then an upper bound of $\Hhat(\vx)$ can be found by inserting Eq.~\eqref{eq:prove_preconditioning_convergence_hessian_bound} into Eq.~\eqref{eq:prove_preconditioning_convergence_quadratic_taylor}.
        \begin{equation}\label{eq:convergence_alternative_upper_bound}
            \Hhat(\vx) \le \Hhat(\vy) + \nabla \Hhat(\vy)^\top (\vx - \vy) + \frac{1}{2} (\vx - \vy)^\top \mathcal{Q}(\vy) (\vx - \vy).
        \end{equation}
        Then, insert $\mathcal{Q} = 2\hat\mQ$ into Eq.~\eqref{eq:convergence_alternative_upper_bound}. Note that $\hat\mQ\coloneqq 2 (\text{diag}(\vq) + \hat{\lambda}\mI)$, so $\hat\mQ\succeq 0$ and $\hat\mQ^\top=\hat\mQ$. Thereafter, the update rule $\vx= \vy - \hat{\mQ}^{-1}\nabla\Hhat(\vy)$ in Eq.~\eqref{eq:rw_update_f2} gives
        \begin{align}
            & \Hhat(\vx) - \Hhat(\vy)\\
            \le& \nabla \Hhat(\vy)^\top (\vx - \vy) + \frac{1}{2} (\vx - \vy)^\top \mathcal{Q}(\vy) (\vx - \vy) \\ 
            = & \nabla \Hhat(\vy)^\top (\vx - \vy) + 2\left( \hat{\mQ}^{\frac{1}{2}} (\vx - \vy) \right)^\top \left( \hat{\mQ}^{\frac{1}{2}} (\vx - \vy)\right)  \\
            = & 2\nabla \Hhat(\vy)^\top \hat{\mQ}^{-1}\nabla \Hhat(\vy) - 2\nabla \Hhat(\vy)^\top (\hat{\mQ}^{-\frac{1}{2}})^\top \hat{\mQ}^{-\frac{1}{2}}\nabla \Hhat(\vy) \\
            = & 0.
        \end{align}
        Therefore, our QN-IRLS in Eq.~\eqref{eq:rw_update_f2} is guaranteed to descend.
    \end{proof}
    
    % \par
    % What we did here is roughly another majorization-minimization procedure. We first found a majorizer of $\Hhat (\vx)$ and then "minimized" it by assigning a preconditioned GD descent direction and picking a step size that guarantees the majorizer to not increase.

\section{Computation Efficiency}
\label{sec:compute_efficiency}
Our RUNG model preserves advantageous efficiency even adopting the quasi-Newton IRLS algorithm. 

\subsection{Time Complexity Analysis}
Each RUNG layer involves computing $\mW$, $\vq$, and the subsequent aggregations. We elaborate on them one by one. We denote the number of feature dimensions $d$, the number of nodes $n$, and the number of edges $m$, which are assumed to satisfy $m\gg 1$, $n \gg 1$ and $d\gg 1$. The number of layers is denoted as $k$. The asymptotic computation complexity is denoted as $\mathcal{O}(\cdot)$.

\paragraph{Computation of $\mW\odot\mA$ and $\mW\odot\tilde\mA$.} $\mW\coloneqq\1_{i\neq j}\frac{d \rho_\gamma(y_{ij})}{d y_{ij}^2}$ is the edge weighting matrix dependent on the node feature matrix $\mF$. The computation of $y_{ij}=\|\frac{\vf_i}{\sqrt{d_i}} -\frac{\vf_j}{\sqrt{\smash[b]{d_j}}}\|_2$ is $\mathcal{O}(\mathpzc{d})$ and that of $\frac{d \rho_\gamma(y_{ij})}{d y_{ij}^2}$ is $\mathcal{O}(1)$. $\mW_{ij}$ only needs computing when $(i,j)\in\mathcal{E}$, because $\forall (i,j)\notin\mathcal{E}$, $\mW_{ij}$ will be masked out by $\mA$ or $\tilde\mA$ anyways. Each element of $\mW$ involves computation time of $\mathcal{O}(d)$ and $m$ elements are needed. In total, $\mW$ costs $\mathcal{O}(md)$, and $\mW\odot\mA$ and $\mW\odot\tilde\mA$ cost $\mathcal{O}(md+m)=\mathcal{O}(md)$.

% \paragraph{Computation of $\vq$.} 
% 
\paragraph{Computation of $\hat\mQ^{-1}$.} $\hat\mQ^{-1} \coloneqq \frac{1}{2}(\text{diag}(\vq^{(k)}) + \lambda \mI)^{-1}$ is the inverse Hessian in our quasi-Newton IRLS. Because $\hat\mQ$ is designed to be a diagonal matrix, its inverse can be evaluated as element-wise reciprocal which is efficient. As for $\vq \coloneqq\sum_{j} \mW^{(k)}_{mj}\mA_{mj} / d_m$, only existing edges $(i,j)\in\mathcal{E}$ need evaluation in the summation. Therefore, this computation costs $\mathcal{O}(m)$. Thus, $\hat\mQ^{-1}$ costs $\mathcal{O}(\mathpzc{m})$ in total.

\paragraph{Computation of aggregation.} An RUNG layer follows 
\begin{equation}
\mF^{(k+1)} = 2\hat\mQ^{-1}\left( ( \mW^{(k)}\odot\tilde{\mA}) \mF^{(k)} +   \lambda \mF^{(0)}\right)  , 
\end{equation}
which combines the quantities calculated above. An extra graph aggregation realized by the matrix multiplication between $\mW\odot\tilde\mA$ and $\mF$ is required, costing $\mathcal{O}(md)$. The subsequent addition to $\mF^{(0)}$ and the multiplication to the diagonal $\hat\mQ^{-1}$ both cost $\mathcal{O}(nd)$.

\paragraph{Stacking layers.} RUNG unrolls the QN-IRLS optimization procedure, which has multiple iterations. Therefore, the convergence increase that QN-IRLS introduces allows a RUNG with fewer layers and increases the overall complexity. It is worth noting that the QN-IRLS utilizes a diagonal approximated Hessian, and thus the computation per iteration is also kept efficient as discussed above.

Summing up all the costs, we have the total computational complexity of our RUNG, $\mathcal{O}((m + n)kd)$. Our RUNG thus scales well to larger graph datasets such as ogbn-arxiv.

\paragraph{Space Complexity Analysis}
The only notable extra storage cost is $\mW$ whose sparse layout takes up $\mathcal{O}(m)$. This is the same order of size as the adjacency matrix itself, thus not impacting the total asymptotic complexity.

\subsection{Alternative Perspective}

In fact, the above analysis can be simplified when we look at the local aggregation behavior of RUNG. For node $i$, it's updated via aggregation $\vf_i = \frac{2}{\hat\mQ_{ii}^{-1}} ((\sum_{j\in\mathcal{N}(i)} \mW_{ij} \vf_j) + \lambda \vf_i^{(0)})$. The summation over neighbors' $\vf_j$ will give $\mathcal{O}(m)$ in the total time complexity in each feature dimension, and $\mW_{ij}$ involves $\mathcal{O}(d)$ computations for each neighbor. This sums up to $\mathcal{O}(md)$ as well. Essentially, the high efficiency of RUNG originates from that every edge weighting in our model involves only the 2 nodes on this edge.

\iffalse
\subsection{Implementation}
The most costly operation is the evaluation of $W$ where $W_{ij}=\rho(\mZ_{ij}) = \rho(\sum_k (\mF_{ik} - \mF_{jk})^2)$. The computation complexity comes to $n^2 d$. However, since $Z_{ij} = \sum_k (F_{ik} - F_{jk})^2 = \sum_k F_{ik}^2 + F_{jk}^2 - 2  F_{ik}  F_{jk}$, the only expensive term is $\sum_k F_{ik}  F_{jk}$. However, efficient implementations have been well studied for matrix multiplications, and using $\sum_k F_{ik}  F_{jk} = (F F^\top)_{ij}$ we can implement the optimized calculation of $\mW$ with ease, without too much worries about memory matrix splitting, etc.
\fi

\section{Comprehensive Bias Analysis}
\label{sec:bias_analysis_compre}
In this section, we provide multiple evidence from various perspectives to reveal the estimation bias of $\ell_1$-based estimation as follows.

(1) From the theoretical perspective, the extensive literature on high-dimensional statistics~\citep{tibshirani1996regression,zhang2010nearly} has proved that $\ell_1$ regularization induces an estimation bias. 

(2) From the algorithm perspective, in Section~\ref{sec:bias_analysis}, we provide the explanation on the $\ell_1$-based estimation problem solver. Specifically, the soft-thresholding operator $S_\lambda(\theta):=\text{sign}(\theta)\max(|\theta|-\lambda, 0) $ induced by the $\ell_1$ regularized problem causes a constant shrinkage for $\theta$ larger than $\lambda$, enforcing the estimator to be biased towards zero with the magnitude $\lambda$.  
% \xr{Zhichao, could you provide the detailed formula of soft-threshold holding to explain the bias?}

(3) From the numerical simulation in Section~\ref{sec:bias_analysis}, we provide an example of mean estimation to verify this estimation bias. As shown in Figure~\ref{fig:m_estimator_compare}, the $\ell_1$ estimator (green) deviates further from the true mean as the ratio of outliers escalates. This can be clearly explained as the effect of the accumulation of estimation bias. In other words, each outlier results in a constant bias, and the bias accumulates with more outliers.
 
(4) From the performance perspective, $\ell_1$-based GNNs such as SoftMedian, TWIRLS, and RUNG-$\ell_1$ (the $\ell_1$ variant of our model) suffer from significant performance degradation when the attack budget increases. 

(5) From our ablation study in Figure 6, we quantify the estimation bias of the aggregated feature $f_i^\star$ on the attacked graph from the feature $f_i$ on the clean graph: $\sumnode \|f_i-f_i^*\|_2^2$. The results demonstrate
that $\ell_1$-based GNN produces biased estimation under adversarial attacks and
the bias indeed scales up with the attack budget. However, our proposed RUNG method exhibits a nearly zero estimation bias under the same attacking budgets.

All of this evidence can convincingly support our claim that $\ell_1$-based robust estimator suffers from the estimation bias, which validates the motivation of our new algorithm design.

\newpage
\section{Additional experiment results}\label{sec:appendix_additional_results}

In this section, we present the experiment results that are not shown in the main paper due to space limits. 
\subsection{Adaptive Attacks}
\autoref{citeseer_adaptive_local_evasion} and \autoref{citeseer_adaptive_global_evasion} are the results of adaptive local and global attacks on Citeseer, referred to in Section~\ref{sec:exp_main}.

\begin{table}[!htb]\small
\centering
\caption{Adaptive local attack on Citeseer. The \textbf{best} and \underline{second} are marked.}
\resizebox{1.0\textwidth}{!}{%
    \begin{tabular}{l c c c c c c}
        \toprule
        \multirow{1}{*}{Model} & \multirow{1}{*}{$0\%$} & \multicolumn{1}{c}{$20\%$} & \multicolumn{1}{c}{$50\%$} & \multicolumn{1}{c}{$100\%$} & \multicolumn{1}{c}{$150\%$} & \multicolumn{1}{c}{$200\%$} \\
        \midrule
        MLP & $69.3 \pm 2.5$ & $69.3 \pm 2.5$ & $\underline{69.3 \pm 2.5}$& $\mathbf{69.3 \pm 2.5}$ &  $\mathbf{69.3 \pm 2.5}$ &  $\mathbf{69.3 \pm 2.5}$ \\
        GCN     & $79.3 \pm 3.3$  & $44.7 \pm 8.8$  & $27.3 \pm 7.7$  & $6.7 \pm 3.7$  & $0.7 \pm 1.3$  & $0.0 \pm 0.0$  \\
        APPNP   & $\mathbf{80.7 \pm 4.4}$  & $50.0 \pm 6.7$  & $39.3 \pm 6.5$  & $16.7 \pm 12.3$  & $16.0 \pm 8.0$  & $0.0 \pm 0.0$  \\
        GAT  & $74.7 \pm 5.0$  & $15.3 \pm 17.5$  & $13.3 \pm 13.5$  & $12.0 \pm 11.3$  & $12.7 \pm 6.5$  & $9.3 \pm 5.3$ \\
        \midrule
        GNNGuard   & $74.7 \pm 4.5$  & $46.0 \pm 10.4$  & $32.7 \pm 11.0$  & $18.0 \pm 7.5$  & $6.0 \pm 3.9$  & $4.0 \pm 3.9$  \\
        RGCN   & $\underline{80.0 \pm 2.1}$  & $46.7 \pm 9.4$  & $32.7 \pm 8.8$  & $10.0 \pm 5.2$  & $0.7 \pm 1.3$  & $0.7 \pm 1.3$ \\
        GRAND  & $77.3 \pm 2.5$  & $56.7 \pm 4.2$  & $44.0 \pm 3.9$  & $16.7 \pm 6.3$  & $0.7 \pm 1.3$ &  $0.0 \pm 0.0$ \\
        ProGNN  & $\underline{80.0 \pm 2.1}$  & $42.7 \pm 7.4$  & $26.0 \pm 5.3$  & $10.0 \pm 4.7$  & $0.7 \pm 1.3$  & $0.0 \pm 0.0$ \\
        Jaccard-GCN  & $78.7 \pm 3.4$  & $46.7 \pm 7.3$  & $28.0 \pm 7.5$  & $6.7 \pm 4.7$  & $0.7 \pm 1.3$  & $0.0 \pm 0.0$ \\
        SoftMedian  & $78.7 \pm 3.4$  & $69.3 \pm 6.5$  & $66.0 \pm 7.1$  & $56.0 \pm 4.4$  & $8.7 \pm 6.9$  & $3.3 \pm 3.0$  \\ 
        TWIRLS   & $77.3 \pm 2.5$  & $69.3 \pm 1.3$  & $68.7 \pm 1.6$  & $57.3 \pm 2.5$  & $36.7 \pm 4.7$  & $26.7 \pm 4.7$ \\
        TWIRLS-T  & $76.0 \pm 3.3$  & $\underline{70.7 \pm 2.5}$  & $68.7 \pm 2.7$  & $62.0 \pm 3.4$  & $52.7 \pm 5.3$  & $47.3 \pm 8.3$ \\
        \midrule
        RUNG-$\ell_1$ (ours) & $\underline{80.0 \pm 3.7}$  & $\mathbf{75.3 \pm 4.5}$  & $\mathbf{73.3 \pm 3.0}$  & $\underline{67.3 \pm 3.3}$  & $36.0 \pm 9.3$  & $26.0 \pm 8.3$ \\ 
        RUNG (ours)  & $77.3 \pm 1.3$  & $\underline{70.7 \pm 5.7}$  & $\underline{69.3 \pm 6.8}$  & $\underline{67.3 \pm 7.1}$  & $\underline{64.0 \pm 5.7}$  & $\underline{61.3 \pm 5.8}$  \\ 
        \bottomrule
    \end{tabular}
}
%  For our model and SoftMedian, the listed accuracy is the lower value between the original model and its soft relaxation.
\label{citeseer_adaptive_local_evasion}
\end{table}

\begin{table}[!htb]\small
\centering
\caption{Adaptive global attack on Citeseer. The \textbf{best} and \underline{second} are marked.}
\resizebox{1.0\textwidth}{!}{%
    \begin{tabular}{l c c c c c c}
        \toprule
        \multirow{1}{*}{Model} & \multirow{1}{*}{Clean} & \multicolumn{1}{c}{$5\%$} & \multicolumn{1}{c}{$10\%$} & \multicolumn{1}{c}{$20\%$} & \multicolumn{1}{c}{$30\%$} & \multicolumn{1}{c}{$40\%$} \\
        \midrule
        MLP & $67.7 \pm 0.3$ & $67.7 \pm 0.3$& $\underline{67.7 \pm 0.3}$& $\mathbf{67.7 \pm 0.3}$& $\mathbf{67.7 \pm 0.3}$ & $\mathbf{67.7 \pm 0.3}$ \\
        GCN      & $74.8 \pm 1.2$  & $66.1 \pm 1.0$  & $60.9 \pm 0.8$  & $53.0 \pm 1.0$  & $47.0 \pm 0.8$ &$41.2\pm1.1$  \\
        APPNP    & $\mathbf{75.3 \pm 1.1}$  & $65.8 \pm 0.9$  & $60.7 \pm 1.6$  & $52.3 \pm 1.6$  & $46.0 \pm 2.0$ & $41.2\pm2.2$\\
        GAT   & $73.4 \pm 1.2$  & $65.4 \pm 1.3$  & $60.4 \pm 1.4$  & $52.6 \pm 2.5$  & $47.2 \pm 3.4$  & $41.2 \pm 4.8$ \\
        \midrule
        GNNGuard & $72.4 \pm 1.1$  & $65.6 \pm 0.9$  & $61.8 \pm 1.4$  & $55.6 \pm 1.4$  & $51.0 \pm 1.3$ &$47.3\pm1.3$ \\
        RGCN     & $74.4 \pm 1.0$  & $66.0 \pm 0.8$  & $60.6 \pm 0.9$  & $52.5 \pm 0.8$  & $46.1 \pm 0.9$ & $40.2\pm1.0$\\
        GRAND    & $74.8 \pm 0.6$  & $66.6 \pm 0.7$  & $61.8 \pm 0.7$  & $53.6 \pm 1.1$  & $47.4 \pm 1.2$ & $42.2\pm0.9$  \\
        ProGNN  & $74.2 \pm 1.3$  & $65.6 \pm 1.1$  & $60.3 \pm 1.1$  & $52.7 \pm 1.4$  & $46.2 \pm 0.9$  & $40.8 \pm 0.6$ \\
        Jaccard-GCN & $74.8 \pm 1.2$  & $66.3 \pm 1.2$  & $60.9 \pm 1.2$  & $53.3 \pm 0.9$  & $46.5 \pm 0.9$  & $41.1 \pm 1.0$ \\
        EvenNet& $74.6 \pm 0.5$  & $66.8 \pm 0.5$  & $62.0 \pm 0.6$  & $55.9 \pm 0.4$  & $51.1 \pm 0.4$  & $47.4 \pm 0.8$ \\
        GARNET& $74.8 \pm 1.3$  & $68.0 \pm 0.9$  & $64.0 \pm 1.1$  & $58.2 \pm 0.7$  & $53.9 \pm 0.8$  & $51.0 \pm 0.9$ \\
        SoftMedian   & $74.6 \pm 0.7$  & $68.0 \pm 0.7$  & $64.4 \pm 0.9$  & $59.3 \pm 1.1$  & $55.2 \pm 2.0$ & $51.9\pm2.1$ \\ 
        TWIRLS & $74.2 \pm 0.8$  & $69.2 \pm 0.8$  & $66.4 \pm 0.7$  & $61.6 \pm 0.9$  & $58.1 \pm 1.2$ & $51.8\pm1.5$\\
        TWIRLS-T   & $73.7 \pm 1.1$  & $69.1 \pm 1.2$  & $66.4 \pm 1.0$  & $62.8 \pm 1.5$  & $60.0 \pm 1.4$  & $57.4 \pm 1.5$ \\ 
        \midrule
        RUNG-$\ell_{1}$ (ours)   & $\mathbf{75.5 \pm 1.1}$  & $\underline{69.3 \pm 1.2}$  & $65.9 \pm 1.2$  & $61.1 \pm 1.1$  & $57.2 \pm 1.4$ & $53.9\pm1.3$  \\ % lam0.8
        RUNG (ours)  & $74.3 \pm 0.7$  & $\mathbf{71.4 \pm 1.0}$  & $\mathbf{69.8 \pm 1.3}$  & $\underline{67.6 \pm 1.2}$  & $\underline{66.5 \pm 1.3}$  & $\underline{65.3 \pm 1.5}$ \\

        \bottomrule
    \end{tabular}
}
%  For our model and SoftMedian, the listed accuracy is the lower value between the original model and its soft relaxation.
\label{citeseer_adaptive_global_evasion}
\end{table}

\subsection{Transfer Attacks}
\label{sec:transfer_attacks}

\begin{figure}[!ht]
\centering
\includegraphics[width=0.85\textwidth]{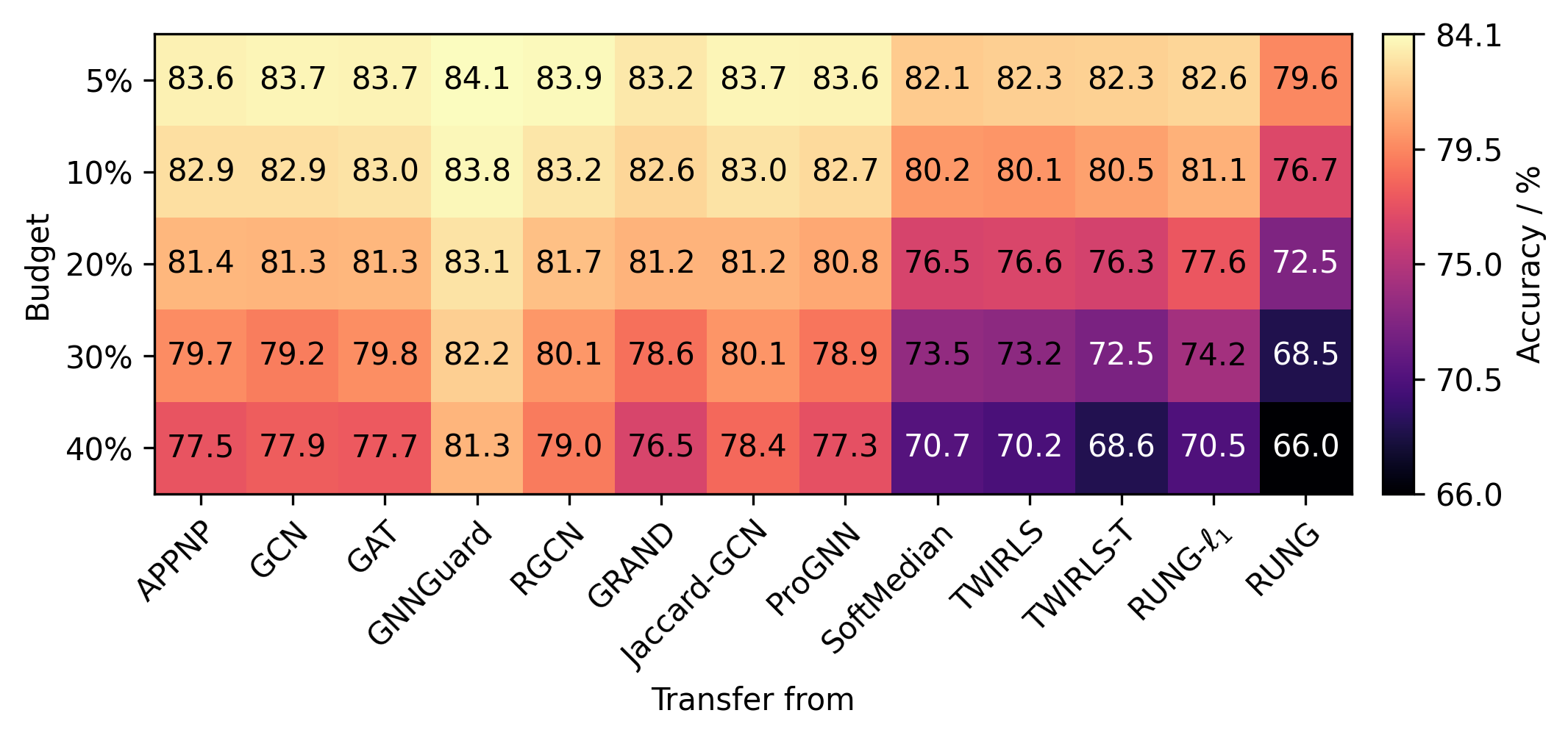}
\vspace{-0.1in}
\caption{Transfer global attack from different surrogate models to our RUNG on Cora ML.
% \xr{we need a legend to show the relation between color and accuracy}\td{if no time to get to this redraw could say something in the caption, eg ``...on Cora, where the darker the color the... ''}
}
\label{fig:cora_transfer_global_evasion_to_mcp}
\vspace{-1ex}
\end{figure}

\begin{figure}[!ht]
    \centering
    \includegraphics[width=0.8\textwidth]
    {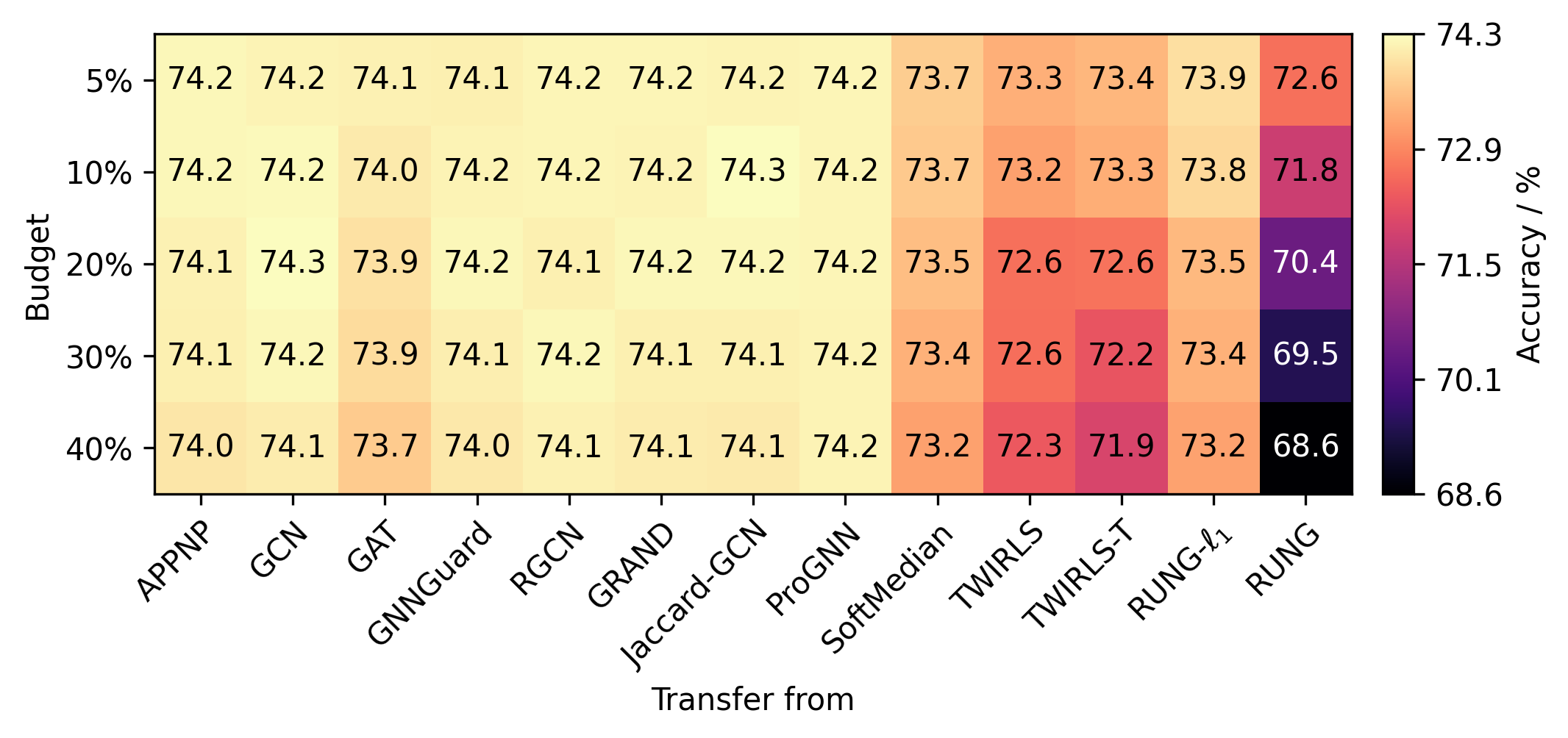}
    \vspace{-0.1in}
    \caption{Transfer global attack from different surrogate models to our RUNG on Citeseer.}
    \label{fig:citeseer_transfer_global_evasion_to_mcp}
\end{figure}

\autoref{cora_transfer_global_evasion} and \autoref{citeseer_transfer_global_evasion} are the results of transfer global attacks on Cora ML and Citeseer. \autoref{fig:cora_transfer_global_evasion_to_mcp} and 
\autoref{fig:citeseer_transfer_global_evasion_to_mcp} are the experiment results of our RUNG attacked by transfer attacks generated on different surrogate models as mentioned in Section~\ref{sec:ablation}.
% \xr{it is better to also provide some description here to tell the reviewer what are these results. }

% \input{tables/cora_adaptive_global_evasion}
% \input{tables/cora_adaptive_local_evasion}

\begin{table}[!ht]\small
\centering
\caption{Transfer global evasion attack on Cora ML. 
}
\resizebox{1\textwidth}{!}{%
    \begin{tabular}{l c c c c c c}
        \toprule
        \multirow{1}{*}{Model} & \multirow{1}{*}{Clean} & \multicolumn{1}{c}{$5\%$} & \multicolumn{1}{c}{$10\%$} & \multicolumn{1}{c}{$20\%$} & \multicolumn{1}{c}{$30\%$} & \multicolumn{1}{c}{$40\%$} \\
                               % & & Adaptive & Transfer       & Adaptive & Transfer       & Adaptive & Transfer       & Adaptive & Transfer \\
        \midrule
        MLP & $65.0 \pm 1.0$ & $65.0 \pm 1.0$ & $65.0 \pm 1.0$ & $65.0 \pm 1.0$ & $65.0 \pm 1.0$& $65.0 \pm 1.0$ \\
        GCN      & $85.0 \pm 0.4$    & $76.0 \pm 0.7$  & $70.6 \pm 0.9$  & $62.1 \pm 1.0$  & $55.4 \pm 1.0$  & $49.8 \pm 1.1$  \\
        APPNP    & $\mathbf{86.3} \pm \mathbf{0.4}$  & $78.0 \pm 1.0$  & $72.7 \pm 1.2$  & $64.4 \pm 1.8$  & $57.7 \pm 1.5$  & $53.0 \pm 1.2$  \\
        GAT  & $83.5 \pm 0.5$ & $78.4 \pm 0.7$  & $75.1 \pm 0.9$  & $69.5 \pm 1.7$  & $65.0 \pm 1.8$  & $61.4 \pm 2.4$ \\
        \midrule
        GNNGuard & $83.1 \pm 0.7$  & $79.9 \pm 0.7$  & $78.1 \pm 0.7$  & $74.8 \pm 0.9$  & $71.8 \pm 1.1$  & $69.9 \pm 1.1$ \\
        RGCN     & $85.7 \pm 0.4$   & $77.7 \pm 0.6$  & $72.7 \pm 0.8$  & $64.4 \pm 0.8$  & $57.9 \pm 0.8$ & $52.6 \pm 1.0$\\
        GRAND    & \underline{$86.1 \pm 0.7$}  & $80.2 \pm 0.7$  & $76.4 \pm 1.0$  & $70.0 \pm 1.3$  & $64.6 \pm 1.4$ & $60.1 \pm 1.1$\\
        ProGNN  & $85.6 \pm 0.5$ & $77.8 \pm 0.7$  & $72.8 \pm 0.9$  & $64.8 \pm 1.1$  & $58.7 \pm 1.3$  & $53.2 \pm 1.3$ \\
        Jaccard-GCN  & $83.7 \pm 0.7$ & $77.8 \pm 0.7$  & $74.1 \pm 1.1$  & $68.4 \pm 1.1$  & $63.5 \pm 1.6$  & $59.3 \pm 1.4$ \\
        %\midrule
        ElasticGNN & $85.9 \pm 0.5$  & \underline{$81.9 \pm 0.9$}  & $79.0 \pm 1.0$  & $73.3 \pm 1.3$  & $68.1 \pm 1.4$ & $63.6 \pm 1.5$\\ 
        SoftMedian  & $85.0 \pm 0.7$  & $81.8 \pm 0.4$  & $79.0 \pm 0.6$  & $73.0 \pm 1.1$  & $66.3 \pm 1.6$ & $60.4 \pm 2.4$ \\        TWIRLS  & $84.2 \pm 0.6$  & $81.5 \pm 0.7$  & \underline{$79.7 \pm 0.6$}  & \underline{$76.8 \pm 0.5$}  & \underline{$74.3 \pm 0.7$} & \underline{$72.8 \pm 0.9$}  \\
        \midrule
        RUNG-$\ell_1$ (ours) & $85.8\pm0.5$ & $82.3 \pm 0.8$  & $79.7 \pm 0.6$  & $75.1 \pm 0.8$  & $70.5 \pm 0.8$  & $67.1 \pm 0.9$ \\
        RUNG (ours) & $84.6 \pm 0.5$  & $\mathbf{83.7} \pm \mathbf{0.6}$  & $\mathbf{82.9} \pm \mathbf{0.9}$  & $\mathbf{81.3} \pm \mathbf{1.3}$  & $\mathbf{79.2} \pm \mathbf{1.6}$ & $\mathbf{77.9} \pm \mathbf{2.0}$ \\ % , Gamma = 3
        \bottomrule
    \end{tabular}
}

\label{cora_transfer_global_evasion}
\end{table}

\begin{table}[!ht]
\centering
\caption{Transfer global evasion attack on Citeseer.}
\resizebox{1\textwidth}{!}{%
    \begin{tabular}{l c c c c c c}
        \toprule
        \multirow{1}{*}{Model} & \multirow{1}{*}{Clean} & \multicolumn{1}{c}{$5\%$} & \multicolumn{1}{c}{$10\%$} & \multicolumn{1}{c}{$20\%$} & \multicolumn{1}{c}{$30\%$} & \multicolumn{1}{c}{$40\%$} \\
        \midrule
        MLP  & $67.7 \pm 0.3$& $67.7 \pm 0.3$& $67.7 \pm 0.3$& $67.7 \pm 0.3$& $\underline{67.7 \pm 0.3}$& $\underline{67.7 \pm 0.3}$\\
        GCN & $74.8 \pm 1.2$  & $66.7 \pm 1.3$  & $62.1 \pm 1.3$  & $54.5 \pm 1.7$  & $48.7 \pm 1.8$ & $43.4 \pm 2.1$ \\
        APPNP   & $\underline{75.3 \pm 1.1}$ & $67.7 \pm 1.2$  & $62.9 \pm 1.0$  & $55.6 \pm 1.0$  & $50.3 \pm 1.0$  & $45.8 \pm 1.6$ \\
        GAT    & $73.4 \pm 1.2$ & $68.2 \pm 1.3$  & $64.6 \pm 1.4$  & $58.2 \pm 2.1$  & $53.2 \pm 3.1$  & $48.7 \pm 3.7$ \\
        \midrule
        GNNGuard  & $72.4 \pm 1.1$ & $70.6 \pm 1.2$  & $69.1 \pm 1.3$  & $67.1 \pm 1.4$  & $65.4 \pm 1.7$  & $64.2 \pm 1.9$ \\
        RGCN  & $74.4 \pm 1.0$  & $67.8 \pm 1.0$  & $63.5 \pm 1.3$  & $56.3 \pm 1.5$  & $51.0 \pm 1.4$  & $46.1 \pm 1.7$ \\
        GRAND  & $74.8 \pm 0.6$  & $68.9 \pm 0.8$  & $65.0 \pm 0.9$  & $59.2 \pm 1.3$  & $54.8 \pm 1.5$  & $51.2 \pm 1.9$ \\
        ProGNN & $74.2 \pm 1.3$ & $66.8 \pm 1.0$  & $62.3 \pm 1.0$  & $55.0 \pm 1.1$  & $49.2 \pm 1.0$  & $43.7 \pm 1.3$ \\
        Jaccard-GCN  & $74.8 \pm 1.2$ & $68.5 \pm 1.1$  & $65.1 \pm 1.0$  & $59.0 \pm 1.7$  & $54.6 \pm 1.9$  & $51.1 \pm 1.7$ \\
        ElasticGNN & ${75.2 \pm 1.2}$  & $70.9 \pm 1.4$  & $67.7 \pm 1.6$  & $62.0 \pm 2.5$  & $57.9 \pm 3.2$   & $53.4 \pm 4.0$ \\ 
        SoftMedian    & $74.6 \pm 0.7$  & $68.0 \pm 0.7$  & $64.4 \pm 0.9$  & $59.3 \pm 1.1$  & $55.2 \pm 2.0$  & $51.9 \pm 2.2$ \\
        TWIRLS & $74.2 \pm 0.8$ & $\underline{71.8 \pm 1.0}$  & $\underline{70.1 \pm 0.9}$  & $\underline{68.4 \pm 1.4}$  & $67.4 \pm 1.6$  & $66.4 \pm 1.8$ \\
        \midrule
        RUNG-$\ell_1$ (ours) & $\mathbf{75.5\pm1.1}$ & $72.0 \pm 1.3$  & $69.3 \pm 1.4$  & $65.1 \pm 1.8$  & $61.8 \pm 2.0$  & $58.7 \pm 2.4$ \\
        RUNG (ours)  & $74.3 \pm 0.7$ & $\mathbf{74.2 \pm 1.0}$  & $\mathbf{74.2 \pm 1.0}$  & $\mathbf{74.3 \pm 1.0}$  & $\mathbf{74.2 \pm 1.1}$  & $\mathbf{74.1 \pm 1.1}$ \\ %Gamma = 1
        \bottomrule
    \end{tabular}
}

%  For our model and SoftMedian, the listed accuracy is the lower value between the original model and its soft relaxation.
\label{citeseer_transfer_global_evasion}
\end{table}

\begin{figure}[!ht]
    \centering
    \includegraphics[width=1.0\textwidth]{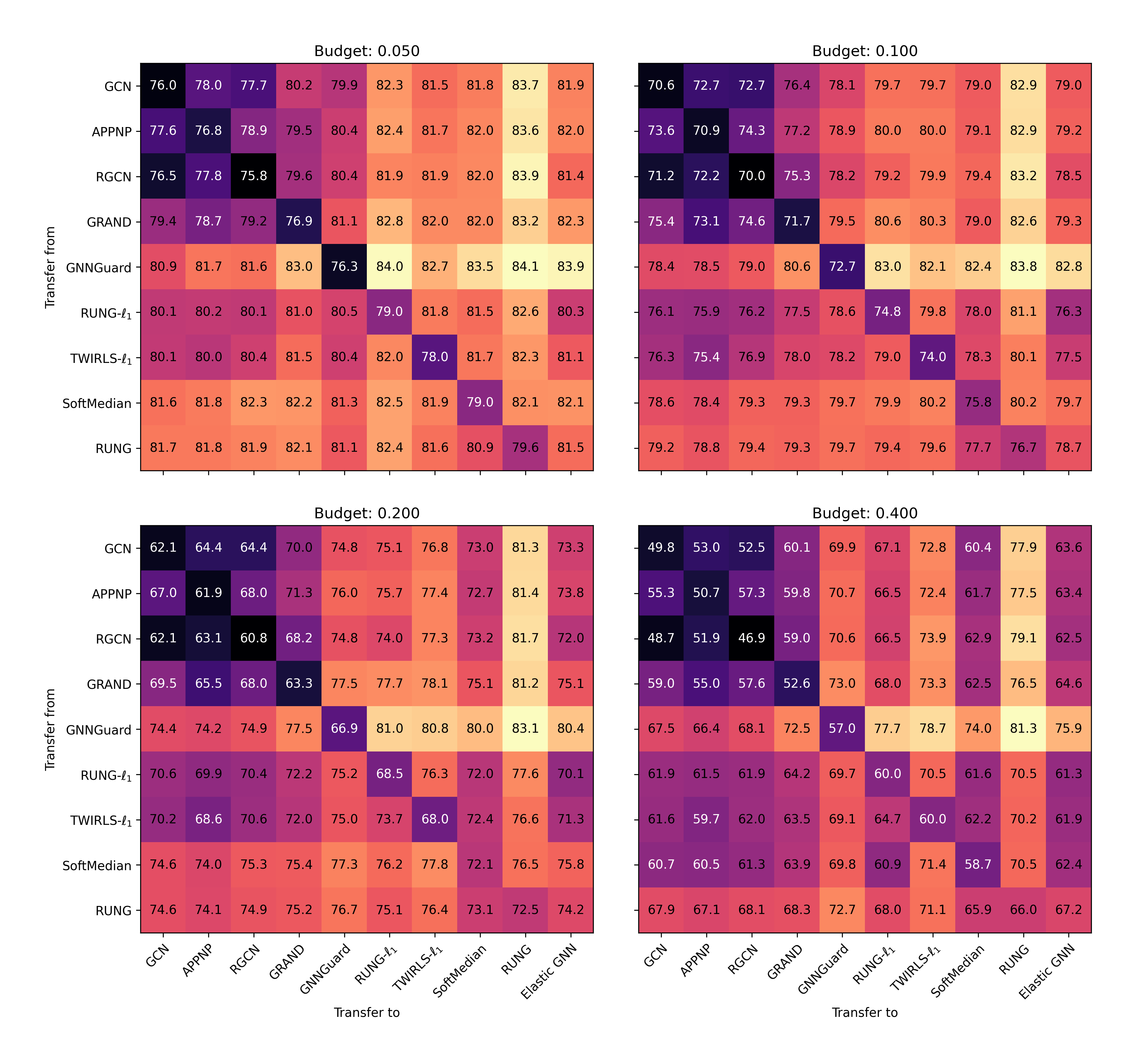}
    \vspace{-0.1in}
    \caption{Transfer global attack between different model pairs on Cora. }
    \label{fig:cora_transfer_all_in_one}
\end{figure}

\autoref{fig:cora_transfer_all_in_one} shows results of global evasion transfer attacks between different models on Cora ML. 
Our observations are summarized below:
\begin{itemize}[leftmargin=*,itemsep=0pt]
    
    \item The attacks generated by RUNG are stronger when applied to more robust models like SoftMedian, while are not strong against undefended or weakly defended models.

    \item For $\ell_1$ GNNs, the attacks are the strongest when transferred from $\ell_1$ GNNs. This supports again our unified view on $\ell_1$ GNNs. An exception is TWIRLS because it only has one attention layer and does not always converge to the actual $\ell_1$ objective.
\end{itemize}

\newpage
\subsection{Poisoning Attacks}
\label{sec:poisoning_attack}

We provide the experiment results under poisoning attacks on  Cora ML and Citeseer in Table~\ref{tab:poisoning_cora} and Table~\ref{tab:poisoning_citeseer}, respectively.

\begin{table}[!ht]
    \centering
    \caption{Poisoning Attacks on Cora ML.}
    
    \begin{tabular}{ccccccc}
    \toprule
        Budget & 5\% & 10\% & 20\% & 30\% & 40\% \\ \hline
        GCN & $74.9 \pm 0.4$ & $69.7 \pm 0.7$ & $60.7 \pm 0.7$ & $54.0 \pm 1.0$ & $48.7 \pm 1.0$ \\ %\hline
        APPNP & $76.3 \pm 0.9$ & $71.1 \pm 1.2$ & $63.0 \pm 1.3$ & $57.1 \pm 0.6$ & $53.2 \pm 1.1$ \\ %\hline
        SoftMedian & $79.2 \pm 0.7$ & $75.6 \pm 0.3$ & $67.8 \pm 0.6$ & $62.9 \pm 1.0$ & $58.6 \pm 0.7$ \\ %\hline
        RUNG-$\ell_1$ (ours) & $79.7 \pm 0.6$ & $76.4 \pm 0.6$ & $68.1 \pm 0.6$ & $63.8 \pm 0.5$ & $60.1 \pm 0.9$ \\ %\hline
        RUNG (ours) & $78.5 \pm 0.5$ & $75.5 \pm 0.3$ & $71.5 \pm 0.4$ & $67.1 \pm 1.6$ & $64.6 \pm 1.3$ \\ \bottomrule
    \end{tabular}

    \label{tab:poisoning_cora}
\end{table}

\begin{table}[!ht]
    \centering
    \caption{Poisoning Attacks on Citeseer.}
    \begin{tabular}{ccccccccccc}
    \toprule
        Budget & 5\% & 10\% & 20\% & 30\% & 40\% \\ \hline
        GCN & $65.5 \pm 1.1$ & $59.8 \pm 1.0$ & $51.0 \pm 1.0$ & $44.0 \pm 1.2$ & $37.9 \pm 1.0$ \\ %\hline
        APPNP & $64.2 \pm 1.8$ & $58.1 \pm 2.6$ & $49.8 \pm 2.5$ & $43.4 \pm 2.3$ & $40.6 \pm 2.7$ \\ %\hline
        SoftMedian & $67.1 \pm 1.0$ & $63.8 \pm 1.0$ & $58.5 \pm 1.1$ & $54.3 \pm 1.9$ & $51.2 \pm 2.4$ \\ %\hline
        RUNG-$\ell_1$ (ours) & $68.9 \pm 1.0$ & $65.9 \pm 1.1$ & $61.0 \pm 1.0$ & $57.2 \pm 1.1$ & $53.9 \pm 1.3$ \\ %\hline
        RUNG (ours) & $72.4 \pm 0.9$ & $72.1 \pm 1.2$ & $71.3 \pm 1.4$ & $70.8 \pm 1.3$ & $69.7 \pm 1.4$ \\ \bottomrule
    \end{tabular}

    \label{tab:poisoning_citeseer}
\end{table}

\subsection{Large Scale Ogbn-Arxiv}
\label{sec:arxiv}

In the large scale datasets, we can not directly apply the vanilla PGD attack~\citep{xu2019topology} on them due to excessive requirement of memory and computation on dense matrix. Alternatively, we leverage the PRBCD~\citep{geisler2021robustness} to scale the PGD attack instead of manipulating on the dense adjacency matrix.
We conduct experiments on large scale Ogbn-Arxiv and the results in~\autoref{tab:global_ogbn_arxiv}  verified the superior robustness of our RUNG model. 
RUNG's robustness outperforms its $\ell_1$ variant which delivers similar performance as SoftMedian and Elastic GNN due to their shared $\ell_1$ graph smoothing.

\begin{table}[!ht]
    \centering
    
    \caption{Global PGD Attacks on Ogbn-Arxiv.}
    \begin{tabular}{cccccc}
    \toprule
        Model & Clean & 1\% & 5\% & 10\% \\ \hline
        GCN & $71.9\pm 0.5$ & $63.1\pm0.4$ & $48.9\pm 3.2$ & $41.8\pm 0.5$ \\ %\hline
        APPNP&$71.7\pm 0.3 $&$64.2\pm 0.4$&$ 50.1\pm 2.3 $& $ 42.2\pm 1.3 $\\
        SoftMedian&$71.2\pm0.5 $&$65.1\pm0.3 $&$54.1\pm 1.6 $& $50.1\pm 2.6 $\\
        RUNG-$\ell_1$ (ours) & $71.6\pm0.6$ & $65.5\pm0.4$ & $55.0\pm 1.1$ & $49.6\pm1.1$ \\ %\hline

        %RUNG ($\gamma = 0.3$) & $64.3\pm1.2$ & $64.2\pm0.2$ & $64.0\pm0.8$ & $61.2\pm0.7$ \\ %\hline
        %RUNG ($\gamma = 1.0$) & $70.2\pm2.1$ & $65.2\pm0.2$ & $55.2\pm0.3$ & $49.3\pm0.1$ \\ %\hline
        RUNG (ours)  & $70.2\pm2.1$ & $65.2\pm0.2$ & $64.0\pm0.8$ & $61.2\pm0.7$ \\ %\hline
        \bottomrule
    \end{tabular}

    \label{tab:global_ogbn_arxiv}

\end{table}

\subsection{Adversarial Training}
\label{sec:adv_train}
We conduct the
adversarial training following~\citep{xu2019topology} and present the results in~\autoref{tab:adv_train}. From the results, we can observe that with adversarial
training, the robustness of RUNG can be further improved.

\begin{table}[!ht]
    \centering
    \caption{Adversarial Training vs Normal Training on RUNG.}
    \resizebox{1.0\linewidth}{!}{
    \begin{tabular}{l|l|l|l|l|l|l}
    \toprule
        Budget & Clean & 5\% & 10\% & 20\% & 30\% & 40\% \\ \hline
        %GCN (Normal Training)     & $85.0 \pm 0.4$  & $75.3 \pm 0.5$  & $69.6 \pm 0.5$  & $60.9 \pm 0.7$  & $54.2 \pm 0.6$ & $48.4\pm0.5$ \\
        %GCN (Adv Training)     & $84.1 \pm 0.2$  & $76.7 \pm 0.3$  & $73.4 \pm 0.7$  & $66.8 \pm 1.2$  & $62.0 \pm 0.8$ & $57.1\pm0.3$ \\
        %\hline
        Normal Training & $84.6 \pm 0.5$  & $78.9 \pm 0.4$  & $75.7 \pm 0.2$  & $71.8 \pm 0.4$  & $67.8 \pm 1.3$ & $65.1 \pm 1.2$\\ % Gamma = 3
        Adversarial Training& $84.3 \pm 1.2$  & $81.8 \pm 0.7$  & $79.9 \pm 0.3$  & $77.3 \pm 1.1$  & $72.8 \pm 0.6$ & $69.1 \pm 0.9$\\ % Gamma = 3
        \bottomrule
    \end{tabular}
    }
    \label{tab:adv_train}
\end{table}

\subsection{Graph Injection Attack}
\label{sec:graph_injection_attack}

The injection attack was
conducted following the settings in~\citep{zou2021tdgia} to evaluate the robustness of different methods. We set up the budget on the
number of injected nodes as 100 and the budget on degree as 200. The results in~\autoref{tab:injection_cite} show that our RUNG significantly
outperforms the baseline models.

\begin{table}[!ht]
\vskip -0.1in
\caption{Graph Injection Attack on Citeseer.}
\setlength\tabcolsep{2.2pt}
\renewcommand{\arraystretch}{1.2}
\vskip 0.0in
\centering
\resizebox{0.4\linewidth}{!}{
\begin{tabular}{c|ccccccccccccccccc}
\toprule
Model&Clean&Attacked\\
\hline
GCN&75.40&28.14\\
APPNP&75.84&25.24\\
GNNGuard&73.47&34.73\\
%GraphCON&74.19&37.29\\
SoftMedian&74.82&38.91\\
RUNG-$\ell_1$&75.01&39.22\\
RUNG-MCP (ours)&75.65&51.13\\
\bottomrule
\end{tabular}
}
\label{tab:injection_cite}
\vskip -0.1in

\end{table}

\newpage
\section{Adaptive Attack Settings}\label{sec:atk_settings}
%\textbf{\color{red} Remove for submission}

For the adaptive PGD attack we adpoted in the experiments, we majorly followed the algorithm in \citep{mujkanovic2022_are_defenses_for_gnns_robust} in the adaptive evasion attack. For the sake of completeness, we describe it below:

In summary, we consider the topology attack setting where the adjacency matrix $\mA$ is perturbed by $\delta\mA$ whose element $\delta\mA_{ij}\in \{0,1\}$. The budget $B$ is defined as $B\ge \|\delta\mA\|_0$. The PGD attack involves first relaxing $\mA$ from binary to continuous so that a gradient ascent attack can be conducted on the relaxed graph.

During the attack, the minimization problem below is solved:
\begin{equation}
    \delta\mA_\star = \argmin_{\delta\mA} \mathcal{L}_{\text{attack}}(\text{GNN}_\theta(\mA+ (\mI-2\mA)\odot\delta\mA, \mF), y_{\text{target}}),
\end{equation}
where $\mathcal{L}$ is carefully designed attack loss function \citep{softmedian,mujkanovic2022_are_defenses_for_gnns_robust}, $\mA$, $\mF$ and $y_{\text{target}}$ are respectively the graph, node feature matrix and ground truth labels in the dataset, $\theta$ is the parameters of the $\text{GNN}$ under attack which are not altered in the evasion attack setting.
$(\mI-2\mA)\odot\delta\mA$ is the calculated perturbation that ``flips'' the adjacency matrix between $0$ and $1$ when it is perturbed.
The gradient of $\frac{\mathcal{L}_{\text{attack}}}{\delta\mA}$ is computed and utilized to update the perturbation matrix $\delta\mA$.

After the optimization problem is solved, $\delta\mA$ is projected back to the feasible domain of $\delta\mA_{ij}\in\{1\}$. The adjacency matrix serves as a probability matrix allowing a Bernoulli sampling of the binary adjacency matrix $\mA'$. The sampling is executed repeatedly so that an $\mA'$ producing the strongest perturbation is finally generated.

\section{Additional Ablation Study of RUNG}
\label{sec:additonal_ablation}

\subsection{Hyperparameters}

\begin{figure}[!ht]
    \centering
    % \includegraphics[width=0.8\textwidth]
    % \begin{adjustwidth}{0.1cm}{0.1cm}
        \includegraphics[width=0.8\textwidth]{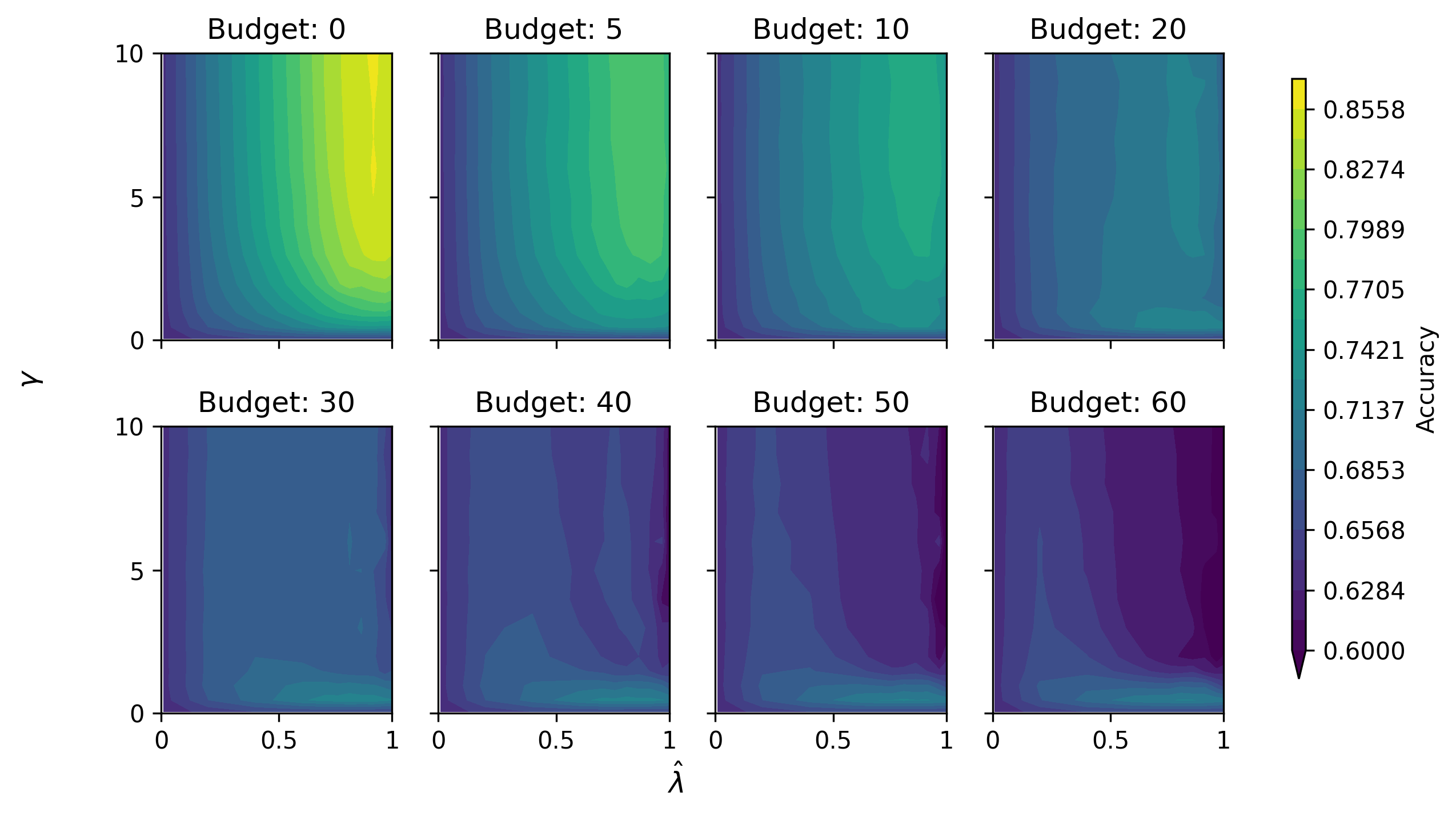}
    % \end{adjustwidth}
    \caption{The performance dependence of RUNG with different hyperparameters $\gamma$ and $\lambda$. The performance is evaluated under different attack budgets. The attack setting is the global evasion attack and the dataset is Cora. Note the x-axis is $\hat\lambda\coloneqq\frac{1}{1+\lambda}$ instead of $\lambda$.}
    \label{fig:tune_mcp}
\end{figure}

The choice of the hyperparameters $\gamma$ and $\lambda$ is crucial to the performance of RUNG. We therefore experimented with their different combinations and conducted adaptive attacks on Cora as shown in Fig. \ref{fig:tune_mcp}. 

Recall the formulation of RUNG in Eq.\eqref{eq:rw_update_f3}:
\begin{equation}
\mF^{(k+1)} = (\text{diag}(\vq^{(k)}) + \lambda \mI)^{-1}\left( ( \mW^{(k)}\odot\tilde{\mA}) \mF^{(k)} +   \lambda \mF^{(0)}\right),
\end{equation}
where $\vq^{(k)}_m = \sum_{j} \mW^{(k)}_{mj}\mA_{mj} / d_m$, $\mW^{(k)}_{ij} = \1_{i\neq j} \max(0, \frac{1}{2\smash[b]{y_{ij}^{\resizebox{!}{3pt}{$(k)$}}}} - \frac{1}{2\gamma})$ and $\smash[b]{y^{(k)}_{ij} = \big\|\frac{\vf_i^{(k)}}{\sqrt{d_i}} - \frac{\vf_j^{(k)}}{\sqrt{\smash[b]{d_j}}}\big\|_2}$. 

In the formulation, $\lambda$ controls the intensity of the regularization in the graph smoothing. In our experiments, we tune $\hat\lambda\coloneqq \frac{1}{1+\lambda}$ which is normalized into $(0,~1)$. In \autoref{fig:tune_mcp}, the optimal value of $\hat\lambda$ can be found almost always near $0.9$ regardless of the attack budget. This indicates that our penalty function $\rho_\gamma$ is decoupled from $\gamma$ which makes the tuning easier, contrary to the commonly used formulation of MCP~\citep{mcp_regression}.

On the other hand, $\gamma$ has a more intricate impact on the performance of RUNG. Generally speaking, the smaller $\gamma$ is, the more edges get pruned, which leads to higher robustness and a lower clean accuracy. We begin our discussion in three cases:

\textbf{Small attack budget ($0\%, 5\%, 10\%$).} The performance is largely dependent on clean accuracy. Besides, when $\gamma\rightarrow\infty$, RUNG becomes a state-of-the-art robust $\ell_1$ model. Therefore, a small $\gamma$ likely introduces more harm to the clean performance than robustness increments over $\ell_1$ models. The optimal $\gamma$ thus at least recovers the performance of $\ell_1$ models.

\textbf{Large attack budget ($20\%, 30\%, 40\%$).} In these cases, $\gamma\rightarrow\infty$ is no longer a good choice because $\ell_1$ models are beginning to suffer from the accumulated bias effect. The optimal $\gamma$ is thus smaller (near $0.5$). However, for fairness, we chose the same $\gamma$ under different budgets in our experiments, so the reported RUNG fixes $\gamma=3$. In reality, however, when we know the possible attack budgets in advance, we can tune $\gamma$ for an even better performance.

\textbf{Very large attack budget ($50\%, 60\%$).} We did not include these scenarios because almost all GNNs perform poorly in this region. However, we believe it can provide some insights into robust graph learning. Under these budgets, more than half of the edges are perturbed. In the context of robust statistics (e.g. mean estimation), the estimator will definitely break down. However, in our problem of graph estimation, the input node features offer extra information allowing us to exploit the graph information even beyond the breakdown point. In the ``peak'' near $(0.9, 0.5)$, RUNG achieves $>70\%$ accuracy which is higher than MLP. This indicates that the edge weighting of RUNG is capable of securely harnessing the graph information even in the existence of strong adversarial attacks. The ``ridge'' near a $\hat\lambda=0.2$, on the other hand, emerges because of MLP. When the regularization dominates, $\lambda\rightarrow\infty$, and $\hat\lambda\rightarrow 0$. A small $\lambda$ is then connected to a larger emphasis on the input node feature prior. Under large attack budgets, MLP delivers relatively good estimation, so a small $\hat\lambda$ is beneficial.

\subsection{GNN Layers}

\begin{figure}[!ht]
    \centering
    % \includegraphics[width=0.8\textwidth]
    % \begin{adjustwidth}{0.1cm}{0.1cm}
        \includegraphics[width=0.5\textwidth]{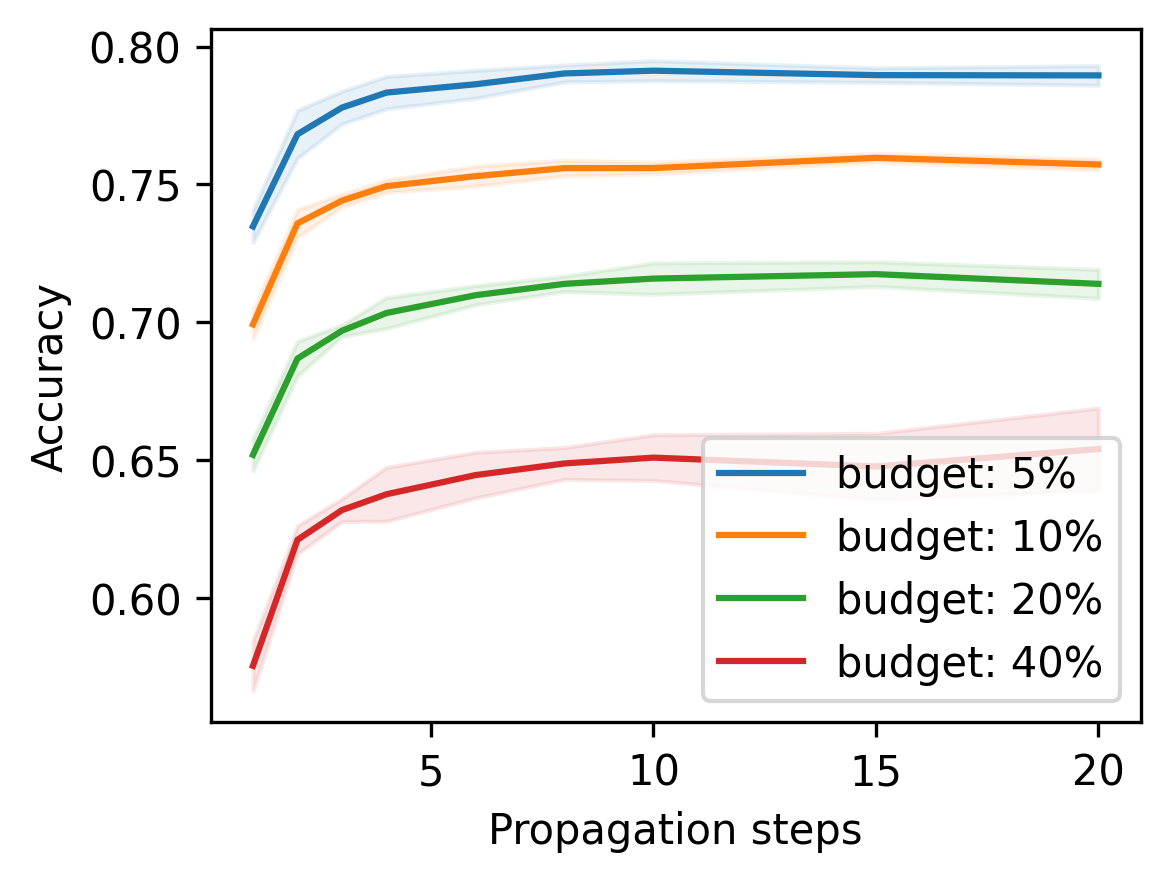}
    % \end{adjustwidth}
    \caption{The performance dependence of RUNG on the number of aggregation layers.}
    \label{fig:layer_number_acc_cora}
\end{figure}

In RUNG, QN-IRLS is unrolled into GNN layers. We would naturally expect RUNG to have enough number of layers so that the estimator converges as desired. We conducted an ablation study on the performance (clean and adversarial) of RUNG with different layer numbers and the results are shown in Fig. \autoref{fig:layer_number_acc_cora}. We make the following observations:
\begin{itemize}[left=0.0pt]
    \item As the layer number increases, RUNG exhibits better performance. This verifies the effectiveness of our proposed RUGE, as well as the stably converging QN-IRLS.
    \item The performance of RUNG can achieve a reasonably good level even with a small layer number ($3$-$5$ layers) with accelerated convergence powered by QN-IRLS. This can further reduce the computation complexity of RUNG.
\end{itemize}

\newpage
\section{Robustness of GCN and APPNP}\label{sec:variants}
%\textbf{\color{red} Remove for submission}

% {\color{blue} \begin{itemize}
%     \item Analysis: GCN allows attacks to "escape" more easily. Linear transforms in conv layers do not necessarily preserve edge distance.
%     \item experiment results: GNNGuard, L1 and MCP; GCN / APPNP
% \end{itemize}}

% Our formulation of the MCP model intrinsically leads to an APPNP-like GNN model, but the reweighting scheme can also be applied to GCN by taking only two steps of the gradient descent during the graph denoising and inserting linear transforms after each of them (see \autoref{sec:variants}). 

In addition to the formulation in \autoref{sec:algo} the main text, we can simply apply our edge reweighting technique to the GCN architecture. Essentially the aggregation operation in GCN can be viewed as an APPNP layer with $\hat{\lambda} = 0$. The GCN version of a layer in our model can then be formulated as 
\begin{equation}
    \mF^{(k+1)} = \text{ReLU} ( (\mW^{(k)} \odot \tilde{\mA})\mF^{(k)} \mM^{(k)}),
\end{equation}
where $\mM$ is the learned weight matrix in GCN.

\paragraph{GCN-based defenses are less robust than APPNP-based defenses}
GCN consists of layers in which both feature transformation and message passing are included. This graph convolution operation will weaken defense methods that rely on edge weighting, such as GNNGuard, models using $l_1$ penalty as well as our method using MCP penalty. 
\par
Consider an edge that is added by the attacker\footnote{Almost all attacks add new edges as shown by our experiments, so this is almost always the case}. A successful defense should detect the attack on this edge by the large difference of nodes connected by this edge, and then assign a zero weight or at least a smaller weight to this edge. However in GCN, even if this edge is detected in the first layer's message passing, the subsequent feature transformation makes the node difference less likely to be preserved until the second layer. This is where the attack can evade the defense and is thus a vulnerability allowing adaptive attacks through. According to our experiments, using different defense parameters in different layers of GCN, unfortunately, does not help much either. 
\par
\begin{figure}[!ht]
    \begin{center}
    %\framebox[4.0in]{$\;$}
    \includegraphics[width=1.0\textwidth]{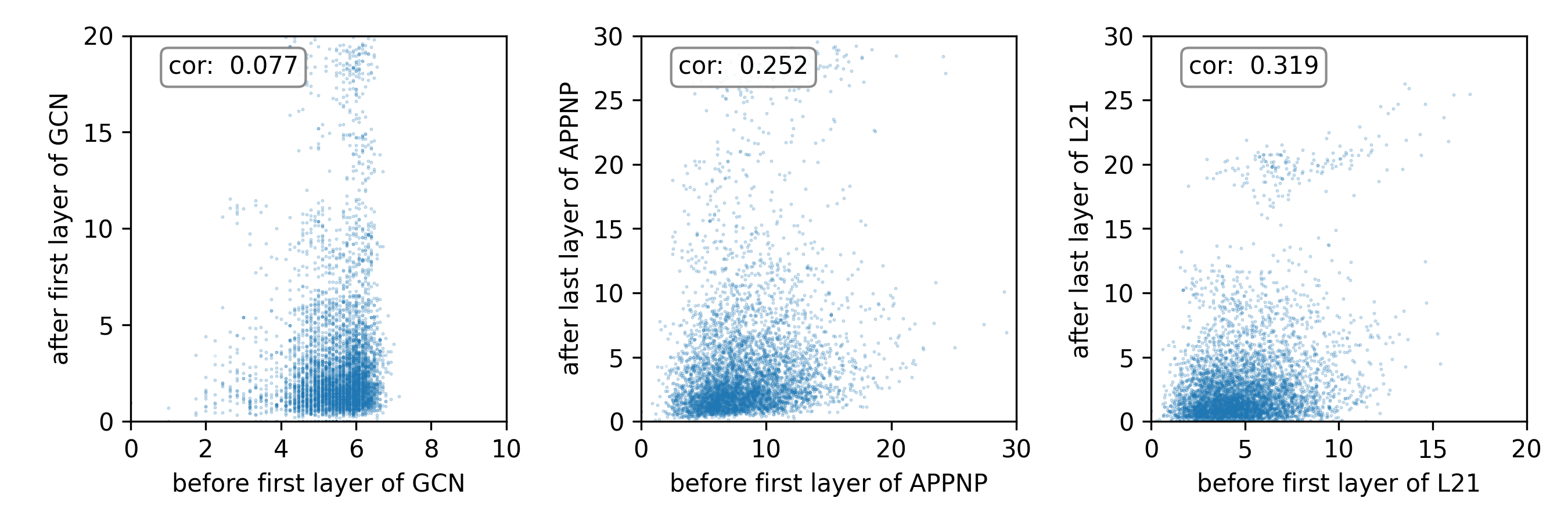}
    \end{center}
    \caption{The distances $\|\vf_i\ - \vf_j\|_2$ between connected nodes in different layers are shown. The linear transform operation in aggregation layers of GCN reduces the correlation between the node distance at different propagation layers.}
    \label{fig:compare_gcn_and_appnp}
\end{figure}

On the other hand, in APPNP node features are also altered along the message-passing layers, but the node distance change is more regulated than in GCN since MLP is decoupled from the graph smoothing layers. In the latter submodule, node differences simply decrease, which allows our defense based on node differences.
\paragraph{Experiments}
It can be seen from \autoref{fig:compare_gcn_and_appnp} that the correlation of node feature differences in different layers of GCN is about 4 times less than in APPNP, which means that an attack detected in the first layer is less likely to continue to be detected in the second layer than in APPNP. This property of GCN makes the many defense methods using GCN architecture as a backbone less robust, as shown in the experiment results in \autoref{cora_adaptive_global_evasion_gcn_vs_appnp} and \autoref{citeseer_adaptive_global_evasion_gcn_vs_appnp}. Nevertheless, our GCN-based MCP model outperforms the SOTA models using $l_1$ methods.

% compare the GCN and APPNP architecture.

\begin{table}[!ht]
\centering
\resizebox{1\textwidth}{!}{%
    \begin{tabular}{l c c c c c c}
        \toprule
        \multirow{1}{*}{Model} & \multirow{1}{*}{Clean} & \multicolumn{1}{c}{$5\%$} & \multicolumn{1}{c}{$10\%$} & \multicolumn{1}{c}{$20\%$} & \multicolumn{1}{c}{$30\%$} & \multicolumn{1}{c}{$40\%$} \\
                               % & & Adaptive & Transfer       & Adaptive & Transfer       & Adaptive & Transfer       & Adaptive & Transfer \\
        \midrule
        RUNG-$\ell_1$ & $85.8 \pm 0.5$  & $78.4 \pm 0.4$  & $74.3 \pm 0.3$  & $68.1 \pm 0.6$  & $63.5 \pm 0.7$ & $59.8\pm0.8$\\ % lambda = 0.8 $l_{21}$-APPNP
        RUNG & $84.6 \pm 0.5$  & $78.9 \pm 0.4$  & $75.7 \pm 0.2$  & $71.8 \pm 0.4$  & $67.8 \pm 1.3$ & $65.1\pm1.2$\\ % Gamma = 3 MCP-APPNP

        % Soft-MCP-APPNP  & $85.6 \pm 0.2$  & $79.0 \pm 0.3$  & $75.5 \pm 0.1$  & $70.4 \pm 0.4$  & $66.8 \pm 0.6$  & $63.6 \pm 1.0$ \\
        % MCP, Gamma = 5: & $85.6 \pm 0.6$  & $79.5 \pm 0.4$  & $76.1 \pm 0.4$  & $71.1 \pm 0.8$  & $67.1 \pm 1.0$ \\

        RUNG-$\ell_1$-GCN & $84.0 \pm 0.4$  & $74.7 \pm 0.6$  & $69.7 \pm 0.7$  & $62.7 \pm 0.6$  & $57.6 \pm 0.6$  & $53.5 \pm 0.8$ \\ % lam = 0.9 Soft-$l_{21}$-GCN 
        RUNG-GCN & $82.6 \pm 0.6$  & $76.1 \pm 0.7$  & $71.0 \pm 0.9$  & $64.3 \pm 1.1$  & $59.9 \pm 0.8$  & $56.5 \pm 1.1$  \\ % Gamma = 0.5, Soft-MCP-GCN 

        \bottomrule
    \end{tabular}
}
\caption{Comparison between APPNP-based and GCN-based QN-IRLS on Cora.}
%  For our model and SoftMedian, the listed accuracy is the lower value between the original model and its soft relaxation.
\label{cora_adaptive_global_evasion_gcn_vs_appnp}
\end{table}

\begin{table}[!ht]
\centering
\resizebox{1\textwidth}{!}{%
    \begin{tabular}{l c c c c c c}
        \toprule
        \multirow{1}{*}{Model} & \multirow{1}{*}{Clean} & \multicolumn{1}{c}{$5\%$} & \multicolumn{1}{c}{$10\%$} & \multicolumn{1}{c}{$20\%$} & \multicolumn{1}{c}{$30\%$} & \multicolumn{1}{c}{$40\%$} \\
                               % & & Adaptive & Transfer       & Adaptive & Transfer       & Adaptive & Transfer       & Adaptive & Transfer \\
        \midrule
        RUNG-$\ell_1$ & $75.5 \pm 1.1$  & $69.3 \pm 1.2$  & $65.9 \pm 1.2$  & $61.1 \pm 1.1$  & $57.2 \pm 1.4$ & $53.9\pm1.3$  \\ % lambda = 0.8 $l_{21}$-APPNP
        RUNG  & $74.3 \pm 0.7$  & $71.4 \pm 1.0$  & $69.8 \pm 1.3$  & $67.6 \pm 1.2$  & $66.5 \pm 1.3$  & $65.3 \pm 1.5$ \\ % MCP-APPNP
        % Soft-MCP-APPNP & $74.3 \pm 0.5$  & $70.3 \pm 0.7$  & $68.2 \pm 0.7$  & $64.4 \pm 2.0$ & $62.0 \pm 2.3$  & $62.1 \pm 0.8$ \\ % Gamma = 1
        
        RUNG-$\ell_1$-GCN & $73.8 \pm 1.2$ & $66.1 \pm 1.1$  & $61.9 \pm 1.0$  & $56.0 \pm 0.7$  & $51.3 \pm 0.8$  & $47.7 \pm 0.6$ \\ % Soft-$l_{21}$-GCN
        RUNG-GCN & $73.0 \pm 0.8$ & $68.2 \pm 0.8$  & $64.3 \pm 1.3$  & $59.4 \pm 1.0$  & $55.7 \pm 1.4$  & $53.1 \pm 1.6$ \\ % gamma = 0.5 Soft-MCP-GCN
        \bottomrule
    \end{tabular}
}
\caption{Comparison between APPNP-based and GCN-based QN-IRLS on Citeseer.}
%  For our model and SoftMedian, the listed accuracy is the lower value between the original model and its soft relaxation.
\label{citeseer_adaptive_global_evasion_gcn_vs_appnp}
\end{table}

%\input{section/appendix/best_defense}
%\input{section/appendix/relation_with_tw}

%exclude for the final version
%\input{tables/heterophily}

% temp
% \input{section/appendix/convergence}
% \input{section/appendix/bias}

\end{document}